\documentclass{article}

\PassOptionsToPackage{round}{natbib}
\usepackage[preprint]{neurips_2026}
\newcommand{\autocastpreprint}{}

\usepackage[utf8]{inputenc}
\usepackage[T1]{fontenc}
\usepackage{hyperref}
\usepackage{url}
\usepackage{booktabs}
\usepackage{tabularx}
\usepackage{amsfonts}
\usepackage{nicefrac}
\usepackage{microtype}
\usepackage{graphicx}
\usepackage{subcaption}
\usepackage{xcolor}
\usepackage{csquotes}
\usepackage{amsmath}
\usepackage{bm}
\usepackage{siunitx}
\usepackage{float}
\usepackage{placeins}

\usepackage[acronym]{glossaries}
\glsdisablehyper
\makeglossaries

\newacronym{afcrps}{$\alpha$fCRPS}{almost fair CRPS}

\newacronym{cln}{CLN}{conditional layer normalisation}
\newacronym{crps}{CRPS}{continuous ranked probability score}

\newacronym{fcrps}{fCRPS}{fair CRPS}
\newacronym{fft}{FFT}{fast Fourier transform}
\newacronym{fno}{FNO}{Fourier neural operator}

\newacronym{ln}{LN}{layer normalisation}

\newacronym{mae}{MAE}{mean absolute error}
\newacronym{mc}{MC}{Monte Carlo}
\newacronym{mse}{MSE}{mean squared error}

\newacronym{pde}{PDE}{partial differential equation}

\newacronym{rmse}{RMSE}{root mean squared error}

\newacronym{uq}{UQ}{uncertainty quantification}

\newacronym{sota}{SotA}{state-of-the-art}
\newacronym{ssr}{SSR}{spread-skill ratio}

\newacronym{vit}{ViT}{vision transformer}
\newacronym{dit}{DiT}{diffusion transformer}
\newacronym{fm}{FM}{flow matching}
\newacronym{vrmse}{VRMSE}{variance scaled root mean squared error}

\newacronym{ad}{AD}{Advection--Diffusion}
\newacronym{gs}{GS}{Gray--Scott}
\newacronym{cns}{CNS}{Conditioned Navier--Stokes}
\newacronym{gpe}{GPE}{Gross--Pitaevskii Equation}
\newacronym{ode}{ODE}{ordinary differential equation}
\newacronym{psrmse}{PSRMSE}{power spectrum RMSE}

\newcommand{\AutoSimRepository}{\url{https://github.com/alan-turing-institute/autosim}}
\newcommand{\AutoCastRepository}{\url{https://github.com/alan-turing-institute/autocast}}
\newcommand{\AutoUQRepository}{\url{https://github.com/alan-turing-institute/autouq}}

\title{Reliability of Probabilistic Emulation \\of Physical Systems}
\author{%
  Sam F. Greenbury\thanks{Equal contribution.}  \\
  The Alan Turing Institute\\
  \texttt{sgreenbury@turing.ac.uk} \\
  \And
  Radka Jersakova\footnotemark[1] \\
  The Alan Turing Institute\\
  \texttt{rjersakova@turing.ac.uk} \\
  \AND
  Paolo Conti \\
  The Alan Turing Institute\\
  Autodesk Research \\
  \And
  Marjan Famili \\
  The Alan Turing Institute\\
  PhysicsX \\
  \And
  Christopher Iliffe Sprague \\
  The Alan Turing Institute\\
  Orbital \\
  \And
  Edwin Brown \\
  The Alan Turing Institute\\
  University of Sheffield
  \And
  Farhan Feroz \\
  The Alan Turing Institute\\
  \And
  Jason D. McEwen \\
  The Alan Turing Institute\\
  University College London \\
  \texttt{jmcewen@turing.ac.uk} \\
}

\begin{document}

\maketitle

\begin{abstract}
    Two dominant approaches have emerged for generating probabilistic forecasts of physical systems: generative models, such as diffusion or flow matching; and ensembles of deterministic models with stochasticity injected, trained using the \gls{crps} loss. While both approaches have demonstrated strong predictive accuracy, the reliability of their uncertainties has not been systematically assessed. We address this gap by developing a framework to evaluate both approaches across diverse 2D spatiotemporal physical systems, under matched model size and computational budget. We assess the reliability of probabilistic emulation by inspecting the empirical coverage of predictive intervals, while also considering accuracy and computational efficiency metrics. \gls{crps}-trained ensembles typically achieve more reliable uncertainties on both single-step prediction and autoregressive rollouts, demonstrating better coverage than the standard alternative of training generative models in a latent space. Moreover, the \gls{crps} approach offers significantly faster inference. When generative models are trained in ambient rather than a compressed latent space, which is often infeasible for high-dimensional problems, they exhibit comparable coverage to \gls{crps}-trained ensembles, though with substantially larger inference latency. In contrast, when \gls{crps}-trained ensembles are trained in latent space they do not show a marked degradation in coverage with respect to ambient space. Both generative models and \gls{crps}-trained ensembles demonstrate good predictive accuracy. To facilitate future research and application, we release \texttt{AutoCast}, a modular framework implementing both generative models and \gls{crps}-trained ensembles, alongside \texttt{AutoSim}, a flexible dataset generation package for rapid prototyping.
\end{abstract}

\section{Introduction}

\glsresetall

The modelling of physical systems underpins discovery across diverse domains from climate to materials science. While computational simulations are often used for modelling these systems, their computational expense bottlenecks their practical deployment. Data-driven emulators offer a fast and accurate alternative. Current benchmarks predominantly focus on deterministic training and point-prediction accuracy \citep{thewell, pdebench, pdearena, ctf4science}. However, this overlooks a critical requirement for reliable \gls{uq} estimates in real-world deployment. Emulators must contend with both \textit{epistemic uncertainty} (arising from the model's incomplete representation of the underlying physics and limited training data) and \textit{aleatoric uncertainty} (arising from inherent stochasticity in the system or noisy observations). Consequently, reliable probabilistic forecasts can be more valuable than marginal improvements in mean prediction error, especially when they inform risk assessment and planning.

Two dominant probabilistic modelling approaches have emerged for physical system emulation: (i) generative models; and (ii) ensembles of deterministic models with stochasticity injected through noise, trained using \gls{crps} loss. Generative models, such as denoising diffusion~\citep{ho2020} and \gls{fm}~\citep{flowmatching}, have demonstrated strong performance, especially for long-term rollouts, but are computationally expensive at inference. To address this, recent work has shown that training in lower-resolution latent spaces can improve efficiency without degrading performance~\citep{lola}. Conversely, \gls{crps}-trained ensembles offer a computationally efficient alternative, even when forecasting directly in ambient space, that has gained significant traction, particularly in weather forecasting applications~\citep{aifscrps, alet2025, fourcastnet}. Both approaches implement losses that target learning the full conditional forecast distribution, with generative models doing so implicitly by targeting a pushforward operator from noise to the target distribution, while \gls{crps}-trained models directly optimise a proper scoring rule through the \gls{crps} loss~\citep{gneiting2007}.

While both approaches have demonstrated strong predictive accuracy, the reliability of their predictive intervals has not been systematically assessed or compared. In practice, miscalibrated uncertainties can result in catastrophic failure of downstream tasks, e.g.\ suboptimal decision making. This motivates a focus on empirical coverage, the empirical proportion $(1-\hat{\alpha})$ of cases in which a $(1-\alpha)$ prediction interval contains the underlying true value. Ideally, the empirical coverage $(1-\hat{\alpha})$ should closely approximate the nominal coverage $(1-\alpha)$. However, achieving well-calibrated uncertainty in probabilistic emulators remains challenging. Recent empirical evaluations indicate that traditional probabilistic methods, such as quantile regression, can exhibit miscalibration in practice \citep{gopakumar2026}, motivating the need for rigorous assessment of the probabilistic reliability of generative and \gls{crps}-trained models.

No work has systematically compared generative models and \gls{crps}-trained ensembles using empirical coverage across diverse physical systems under controlled experimental conditions. Because both approaches target the full conditional forecast distribution, one might expect them to yield reliable predictive uncertainties. In practice, however, each approach relies on approximations, so empirical study is needed to determine how reliable their probabilistic forecasts are. We address this gap by assessing \gls{uq} reliability via empirical coverage across a number of 2D spatiotemporal physical systems. Our findings highlight promising future research directions to ensure reliable \gls{uq} for complex physical emulation, a necessary requirement for real-world deployment.

The remainder of this article is structured as follows. Section~\ref{sec:simulations} introduces \texttt{AutoSim}, a lightweight dataset generation package for flexible benchmarking, and the studied 2D spatiotemporal systems. Section~\ref{sec:methodology} presents our methodology, describing the two probabilistic frameworks (generative modelling and \gls{crps}-trained ensembles) alongside \texttt{AutoCast}, a modular modelling and benchmarking framework. Numerical experiments and results are outlined in Section~\ref{sec:experiments}, comparing probabilistic reliability, accuracy, and computational cost of the two frameworks. Concluding remarks and future research directions motivated by this work are discussed in Section~\ref{sec:conclusions}.

\section{Simulations}
\label{sec:simulations}

\subsection{AutoSim}

We developed \texttt{AutoSim}\footnote{\AutoSimRepository}, an open-source Python package for flexible dataset generation, to complement the existing benchmarking ecosystem for physical system emulation. 

Existing datasets, such as The Well~\citep{thewell}, PDEBench~\citep{pdebench}, or BubbleML~\citep{bubbleml}, provide high-quality simulation data. However, their high-resolution nature is not always well suited to rapid experimentation or constrained compute budgets. \texttt{AutoSim} addresses this by providing multiple simulators through a consistent API, enabling users to generate datasets at the resolution, simulation horizon and trajectory count suitable to their compute.\footnote{The primary benefit of \texttt{AutoSim} is that researchers can generate data from multiple physical systems in a consistent format without having to become familiar with specialised simulation software, making ML benchmarking more accessible. That said, \texttt{AutoSim} is not a general-purpose simulation package. Frameworks such as COMSOL~\citep{comsol}, FEniCS~\citep{fenics}, PhiFlow~\citep{phiflow}, and Lily Pad~\citep{lilypad} provide more flexible simulation capabilities.}

\begin{figure}[h]
    \centering
    \includegraphics[width=\linewidth]{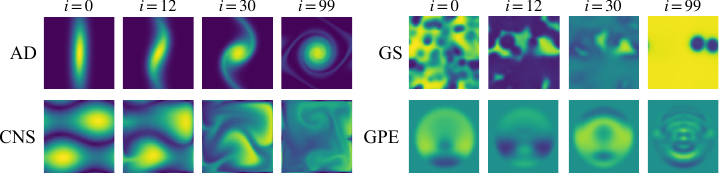}
    \caption{Snapshots at $i=\{0,12,30,99\}$ of the \gls{ad} and \gls{cns} datasets (left) and the \gls{gs} and \gls{gpe} datasets (right); within each dataset, columns are ordered left-to-right by increasing index $i$.}
    \label{fig:datasets_overview}
\end{figure}

We focus on 2D spatiotemporal problems simulated on a regular $64\times 64$ spatial grid with uniform time-step. For this paper, we selected a subset of the simulations available in \texttt{AutoSim}, chosen to demonstrate applicability across different dynamical regimes and levels of complexity. We give a short description of each physical system alongside a visualisation (see Figure~\ref{fig:datasets_overview}). 

\textbf{Advection--Diffusion (AD)} models the evolution of a single vorticity field under nonlinear advection and diffusion. We use it as a simple baseline: it provides a controlled single-channel transport problem that isolates the need to preserve coherent structures while tracking dissipative smoothing.

\textbf{Gray--Scott (GS)} is a two-species reaction--diffusion system whose behaviour varies markedly with the feed and kill rates, producing qualitatively different regimes such as spots, worms, spirals, and maze-like patterns. It tests whether models can represent qualitatively different morphologies within a single \gls{pde} family.

\textbf{Conditioned Navier--Stokes (CNS)} couples passive-scalar transport to an incompressible buoyancy-driven flow. We include it to represent turbulence-like classical fluid behaviour with controllable forcing and boundary-condition variants.

\textbf{Gross--Pitaevskii Equation (GPE)} is a nonlinear quantum-fluid model of a Bose--Einstein condensate driven by a moving laser obstacle, spanning diverse superfluid phenomena including quantum turbulence. It extends the benchmark beyond classical fluids.

For each physical system, we generate multiple deterministic trajectories by varying initial conditions and simulation parameters to capture a diverse range of system dynamics. Specifications of each dataset, including governing equations and numerical methods, are provided in Appendix~\ref{appendix:data_details}.

\section{Methodology}
\label{sec:methodology}

\begin{figure}[t]
    \centering
    \includegraphics[width=\linewidth]{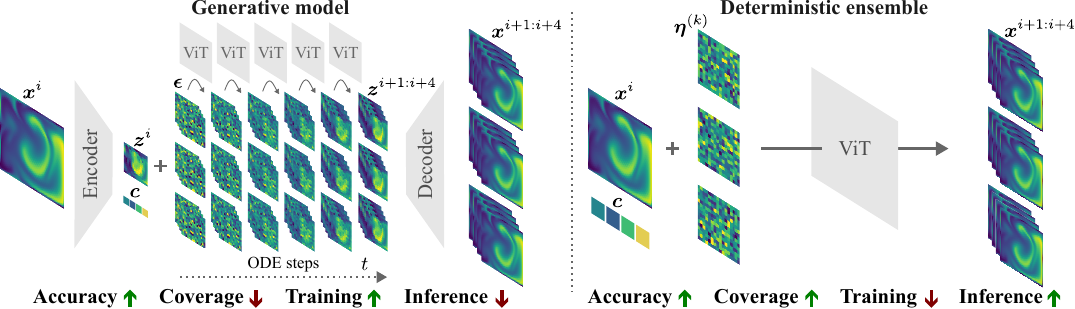}
    \caption{Illustration of the two inference pipelines: generative model and deterministic ensemble trained with CRPS. Both processors predict an output window of $n=4$ future states given an input window of $m=1$ state. For illustrative purposes we show an ensemble size of $M=3$. For generative models, prediction happens in the latent rather than ambient space. Arrows indicate relative advantages and disadvantages. CRPS training is slower per epoch but it yields better coverage and faster inference than the generative model. The two approaches have comparable predictive accuracy.}
    \label{fig:methods}
\end{figure}

This section defines the probabilistic forecasting setup and the two modelling pipelines compared throughout the paper (see Figure \ref{fig:methods}). We first introduce the windowed single-step prediction and autoregressive rollout tasks, then describe latent space \gls{fm} and ambient space \gls{crps}-trained ensembles as alternative ways to represent forecast distributions. We then review scoring rules and coverage diagnostics used to assess probabilistic reliability before summarising the \texttt{AutoCast} implementation used in the experiments.

\subsection{Spatiotemporal forecasting}

The goal of spatiotemporal forecasting is to predict a future sequence of states $\bm{x}^{i+1:i+n}$ given a present sequence $\bm{x}^{i-m+1:i}$ and any additional conditioning on simulation parameters $\bm{c}$, which we treat here as constant over the sequence. We write the input window as $\bm{x}_{\mathrm{in}}=\bm{x}^{i-m+1:i}\in\mathbb{R}^{m\times W\times H\times C}$ and the output window as $\bm{x}_{\mathrm{out}}=\bm{x}^{i+1:i+n}\in\mathbb{R}^{n\times W\times H\times C}$, where $m$ and $n$ are, respectively, the input and output window sizes, $W$ and $H$ the spatial width and height, and $C$ the number of channels. For probabilistic forecasting, we aim to learn the conditional distribution $p(\bm{x}_{\mathrm{out}}\mid\bm{x}_{\mathrm{in}}, \bm{c})$ and train models that represent the approximation $p_\theta (\bm{x}_{\mathrm{out}} | \bm{x}_{\mathrm{in}}, \bm{c})$, parameterised by $\theta$, from which we can draw samples. The input window can either be the ground truth (i.e., a single-step prediction) or the model's previous prediction (autoregressive rollout)~\citep{fourcastnet, pdearena}. For latent space models, an encoder $e_\psi$ and decoder $d_\varphi$, with parameters $\psi$ and $\varphi$ respectively, are trained to target $\hat{\bm{x}}^i = d_\varphi(e_\psi(\bm{x}^i))$, defining latent states $\bm{z}^i=e_\psi(\bm{x}^i)$ and latent windows $\bm{z}_{\mathrm{in}}=\bm{z}^{i-m+1:i}$ and $\bm{z}_{\mathrm{out}}=\bm{z}^{i+1:i+n}$. Autoregressive rollout recursively feeds predictions back as inputs with stride $n$; with $m=1$, the first prediction conditions on the ground-truth initial state $\bm{x}^0$,
\[\hat{\bm{x}}^{1:n}\sim p_\theta(\cdot\mid\bm{x}^0,\bm{c}),\qquad
    \hat{\bm{x}}^{i+1:i+n}\sim p_\theta(\cdot\mid\hat{\bm{x}}^i,\bm{c})
    \quad\text{for }i=n,2n,\ldots\]
For latent models, the initial window is encoded once, $\hat{\bm{z}}^{0:m-1}=e_\psi(\bm{x}^{0:m-1})$, after which the recursion is performed entirely in latent space; predictions are decoded into ambient space with $d_\varphi$.

\subsection{Model uncertainty}

The simulations are deterministic, the current state and parameters $\bm{c}$ uniquely determine the future trajectory, so the ideal conditional target is the point mass $p(\bm{x}_{\mathrm{out}}\mid\bm{x}_{\mathrm{in}},\bm{c})=\delta\!\left(\bm{x}_{\mathrm{out}}-f_{\mathrm{sim}}(\bm{x}_{\mathrm{in}},\bm{c})\right)$, where $f_{\mathrm{sim}}$ is the simulator map to the output window and $\delta$ denotes the Dirac delta. Nevertheless, epistemic uncertainty remains in the learned dynamics due to finite training data and limited model capacity.\footnote{For the \gls{gpe} dataset, absolute simulation time, and therefore the phase $\omega t$ of the sinusoidal laser motion (Eq.~\eqref{eq: GPE_obs}), is not supplied to the emulator. To the extent that this phase cannot be inferred from the observed state, some residual aleatoric uncertainty might be present.} The trained models output predictive intervals that represent the overall learned uncertainty without distinguishing between epistemic and aleatoric components.

\subsection{Generative models}\label{methods:generative_models}
\label{sec:methodology:generative}

Generative models learn to sample from the conditional forecast distribution by transforming noise from a base distribution, conditioned on the input window and simulation parameters. Diffusion \citep{ho2020} and flow matching (FM) \citep{flowmatching} are common examples. Iterative generative sampling can be computationally costly for large spatial fields, motivating generation in a learned latent space \citep{rombach2022, lola}. Recent work has also found latent diffusion models to improve accuracy and rollout stability of physics emulators relative to deterministic alternatives \citep{lola}. We focus on \gls{fm} because it can offer faster sampling than diffusion \citep{flowmatching}. We apply \gls{fm} in latent space: the encoder $e_\psi$ is applied to each frame of $\bm{x}_{\mathrm{in}}$ and $\bm{x}_{\mathrm{out}}$ to form latent windows $\bm{z}_{\mathrm{in}}$ and $\bm{z}_{\mathrm{out}}$, samples are generated from $p_\theta(\bm{z}_{\mathrm{out}}\mid\bm{z}_{\mathrm{in}},\bm{c})$, and the decoder $d_\varphi$ is applied to each predicted latent frame to obtain $\hat{\bm{x}}_{\mathrm{out}}$.

\gls{fm} defines a conditional target probability path from base noise $p_0=\mathcal{N}(0,\bm{I})$ to the latent forecast distribution $p_1=p(\bm{z}_{\mathrm{out}}\mid\bm{z}_{\mathrm{in}},\bm{c})$. At sampling time, the learned flow map $\phi_{\theta,t}$ converts a noise sample $\bm{\epsilon}\sim p_0$ to $\bm{z}_t=\phi_{\theta,t}(\bm{\epsilon})$, with model pushforward $\hat{p}_t=\left[\phi_{\theta,t}\right]_\#p_0$, and is represented through a velocity field $v_\theta$ via the \gls{ode} form
\[\frac{d\bm{z}_t}{dt}=v_\theta(\bm{z}_t,t,\bm{z}_{\mathrm{in}},\bm{c}),\qquad
    \bm{z}_1=\bm{\epsilon}+\int_0^1 v_\theta(\bm{z}_t,t,\bm{z}_{\mathrm{in}},\bm{c})\,dt.\]
Following \citet{flowmatching}, for the linear interpolation path, $\bm{z}_t=(1-t)\bm{\epsilon}+t\bm{z}_{\mathrm{out}}$ and $u_t(\bm{z}_t\mid\bm{\epsilon},\bm{z}_{\mathrm{out}})=d\bm{z}_t/dt=\bm{z}_{\mathrm{out}}-\bm{\epsilon}$, the velocity field is trained with loss
\[\mathcal{L}_{\mathrm{FM}}(\theta)=
    \mathbb{E}_{t,(\bm{z}_{\mathrm{in}},\bm{c},\bm{z}_{\mathrm{out}}),\bm{\epsilon}}
    \left[\left\|v_\theta(\bm{z}_t,t,\bm{z}_{\mathrm{in}},\bm{c})
    -u_t(\bm{z}_t\mid\bm{\epsilon},\bm{z}_{\mathrm{out}})\right\|^2\right].\]

\subsection{Deterministic ensembles with noise injection}
\label{sec:methodology:ensemble}

Efficient deterministic-backbone ensembles are widely used for probabilistic weather forecasting \citep{aifscrps, alet2025, fourcastnet}. Deterministic ensembles produce probabilistic forecasts by drawing independent noise variables $\bm{\eta}^{(k)}\in\mathbb{R}^{d_\eta}$ with $\bm{\eta}^{(k)}\sim\mathcal{N}(0,\bm{I})$, where $d_\eta=n_{\mathrm{noise\_channels}}$, and evaluating $\hat{\bm{x}}_{\mathrm{out}}^{(k)}=f_\rho(\bm{x}_{\mathrm{in}},\bm{c},\bm{\eta}^{(k)})$ for processor model $f_\rho$ with parameters $\rho$, where ensemble members are trained jointly using a \gls{crps}-family loss. $\bm{\eta}$ is typically injected into the deterministic backbone using a conditional normalisation strategy~\citep{aifscrps, alet2025, retrofitting} (further implementation details are provided in Appendix~\ref{appendix:model_archs}).

A \gls{crps}-family ensemble objective compares the forecast distribution with the realised outcome. For deterministic forecasts, \gls{crps} reduces to mean absolute error; in the ensemble form below, it becomes the error term minus a pairwise member-distance term, encouraging forecasts that are both accurate and appropriately dispersed. For an ensemble $\{\hat{\bm{x}}_{\mathrm{out}}^{(k)}\}_{k=1}^M$ and target $\bm{x}_{\mathrm{out}}$, with $\|\cdot\|_1$ denoting mean absolute difference over output dimensions, the loss is given by
\[\begin{aligned}
        \mathcal{L}_\lambda
         & = \frac{1}{M}\sum_{j=1}^M
        \bigl\|\hat{\bm{x}}_{\mathrm{out}}^{(j)}-\bm{x}_{\mathrm{out}}\bigr\|_1
        - \lambda\sum_{j=1}^M\sum_{k=1}^M
        \bigl\|\hat{\bm{x}}_{\mathrm{out}}^{(j)}-\hat{\bm{x}}_{\mathrm{out}}^{(k)}\bigr\|_1 .
    \end{aligned}\]
In practice, finite-ensemble variants such as \gls{fcrps} \citep{alet2025} or \gls{afcrps} \citep{aifscrps}, with $\alpha=1$ recovering \gls{fcrps}, are typically used. Ensemble models trained this way are a useful alternative to generative models because each ensemble member requires a single forward pass, making inference much faster. In $\mathcal{L}_\lambda$, the \gls{crps} variants use
\[\begin{aligned}
        \lambda_{\mathrm{CRPS}} & = \frac{1}{2M^2},\qquad
        \lambda_{\mathrm{fCRPS}} = \frac{1}{2M(M-1)},\qquad
        \lambda_{\mathrm{afCRPS}_\alpha} = \frac{M-1+\alpha}{2M^2(M-1)} .
    \end{aligned}\]

\subsection{Proper scoring rules}\label{section:proper_scoring_rules}

\textit{Scoring rules} are used to evaluate the quality of probabilistic forecasts, assigning a numerical score $S(\hat{p}, \bm{x})$ to a sample $\bm{x}$ given predictive distribution $\hat{p}$~\citep{gneiting2007}. A scoring rule is \textit{proper} if its expected value is minimised when the predictive distribution $\hat{p}$ matches the true underlying data distribution $p$, i.e.
\[\mathbb{E}_{\bm{x} \sim p}[S(p, \bm{x})] \le \mathbb{E}_{\bm{x} \sim p}[S(\hat{p}, \bm{x})].\]
A scoring rule is \textit{strictly proper} if equality holds only for $\hat{p}=p$, i.e., only when the predictive distribution matches the true underlying data distribution.

The \gls{crps}-family losses are proper scoring rules for scalar predictive distributions. Averaged componentwise across time, space, and channels, they are proper for the marginals rather than the joint distribution. In contrast, \gls{fm} implicitly optimises the predictive distribution toward the conditional data distribution by learning a flow map whose pushforward distribution $\left[\phi_{\theta,1}\right]_\#p_0$ is designed to match it. Since both \gls{crps}-training and \gls{fm} target the conditional data distribution (componentwise for \gls{crps}, joint for \gls{fm}), both approaches are theoretically expected to produce reliable learned distributions. In practice, there are approximations and sources of error introduced by each approach. For \gls{crps}-trained models, the loss is computed from a fixed, typically small, number of ensemble members. For \gls{fm}, discretisation error is introduced in the \gls{ode} solver, accumulating drift along the trajectory, and distributions are matched in the latent rather than the ambient space. Given both approaches are theoretically well-motivated, empirical studies, such as presented in this article, are necessary to assess how internal errors and approximations impact the overall reliability of probabilistic forecasts. These findings can, in turn, inform future research aimed at mitigating the limitations of current approaches.

Beyond \gls{crps}, other proper scores can target different aspects of the predictive distribution. Multivariate scores, such as the energy score, can capture joint structure across output dimensions~\citep{gneiting2007, pic2024}. Proper scoring rules can also be constructed for cases where forecasts are summarised by intervals: the Winkler interval score,
\[W_{\alpha}(l,u;y)=(u-l)+\frac{2}{\alpha}(l-y)\mathbf{1}\{y<l\}+\frac{2}{\alpha}(y-u)\mathbf{1}\{y>u\},\]
targets both coverage and interval sharpness by penalising wide intervals and missed coverage, and so directly relates to the empirical-coverage diagnostics introduced next.

\subsection{Reliability of learned distributions: coverage testing}\label{sec:coverage_testing}

Probabilistic emulator studies that report \gls{uq} quality metrics often use consistency metrics such as \gls{ssr}, which compares ensemble spread to prediction errors~\citep{lola, alet2025}. Self-consistency as measured by SSR$=1$ is important but not a sufficient condition for reliable forecasts \citep{retrofitting, lola}. \gls{ssr} is limited by the fact that it does not provide a direct measure of whether predictive intervals contain the true value across quantiles of the distribution. We focus here instead on empirical coverage, which tests whether the true output window falls inside prediction intervals at the expected rate. For an ensemble $\{\hat{\bm{x}}_{\mathrm{out}}^{(k)}\}_{k=1}^M$, let $l_{j,r}^{(\alpha)}$ and $u_{j,r}^{(\alpha)}$ be the $\alpha/2$ and $1-\alpha/2$ quantiles for output component $r$ of test case $j$, and let $(\bm{x}_{\mathrm{out}}^{(j)})_r$ be the corresponding truth; then empirical coverage is computed by
\[1-\hat{\alpha}=\frac{1}{NR}\sum_{j=1}^{N}\sum_{r=1}^{R}
    \mathbf{1}\Bigl\{(\bm{x}_{\mathrm{out}}^{(j)})_r
    \in\bigl[l_{j,r}^{(\alpha)},u_{j,r}^{(\alpha)}\bigr]\Bigr\}.\]
Here $r$ indexes all output time, spatial, and channel dimensions of $\bm{x}_{\mathrm{out}}$. For a well-calibrated model, $1-\hat{\alpha}$ should approximate $1-\alpha$ across coverage levels.

\subsection{AutoCast}

As part of this work, we release \texttt{AutoCast}\footnote{\AutoCastRepository}, a flexible Python package for spatiotemporal forecasting that supports training in both ambient and latent space through an encoder--processor--decoder architecture. Its modular design makes it easy to swap out different processors or autoencoders within the same training pipeline.  The package implements \gls{sota} generative models (diffusion, \gls{fm}) alongside popular deterministic architectures (\gls{vit}: \citealt{vit}, U-Net: \citealt{unet}, Fourier neural operators: \citealt{fno}). \texttt{AutoCast} also enables training ensembles with \gls{crps} loss using a variety of noise injection strategies, e.g.\ adaptive layer normalisation (as presented here) or concatenating noise on the inputs.

\texttt{AutoCast} provides out-of-the-box support for \texttt{AutoSim} data, and we additionally target interoperability with \texttt{The Well} dataset~\citep{thewell} given its scale, breadth, and popularity: \texttt{AutoCast} provides data loaders for both \texttt{AutoSim} and \texttt{The Well}, making it easy to switch between them in the same training pipeline and go from rapid prototyping to benchmarking models at scale. Beyond these defaults, \texttt{AutoCast} is extendable to alternative approaches and systems through common interfaces.

\setcounter{topnumber}{2}
\section{Numerical experiments and results}
\label{sec:experiments}

We focus on assessing and comparing the two training pipelines; namely, generative modelling and deterministic ensembles trained with \gls{crps} loss. We consider the reliability of probabilistic forecasts, as assessed by coverage testing, forecast accuracy and computational cost. First we describe the experimental setup, followed by the main comparison, before presenting ablation studies.

\subsection{Experimental setup}\label{sec:experimental_setup}

All training is carried out in \texttt{AutoCast}. All models are trained to predict an output window of $n=4$ future steps from a ground-truth input window of $m=1$, conditioned on any simulator parameters (i.e., single-step prediction). During validation and testing the models are additionally evaluated on autoregressive rollouts for $100$ steps from an initial state and simulator parameters, which lets us measure how accuracy and uncertainty calibration evolve with lead time. For table summaries, we report coverage MAE as the mean absolute difference between empirical and nominal coverage over the evaluated coverage levels.

We consider the two pipelines outlined in Section~\ref{sec:methodology:generative} and Section~\ref{sec:methodology:ensemble} (Figure~\ref{fig:methods}): (i) a generative pipeline in latent space, where an autoencoder is trained first and then frozen, with the \gls{fm} processor trained in latent space; and (ii) an ambient space pipeline with an ensemble processor trained with the \gls{afcrps} loss. The \gls{crps} model uses $M=8$ ensemble members and \gls{fm} sampling uses $50$ Euler steps. To keep the comparison fair, we fix the processor budget: both processors use a \gls{vit} backbone, are constrained to roughly $80$M parameters and are trained under the same maximum wall-clock budget of $4$ GPUs for $23.5$ hours ($94$ GPU-hours). All experiments were run on NVIDIA GH200 Grace Hopper GPUs on the Isambard-AI facility~\citep{isambard}.

\gls{crps} is only strictly proper for scalar marginals, and its componentwise and finite-ensemble approximations have known limitations on multivariate spatial forecasts that have motivated both auxiliary spectral losses~\citep{fourcastnet} and finite-ensemble corrections such as \gls{afcrps}~\citep{aifscrps}, which we adopt for \gls{crps} training. We do not add further regularisation, but observed in some runs that interval coverage degraded while validation loss kept falling, consistent with drift toward an MAE-like regime. We therefore select \gls{crps} checkpoints using the validation Winkler score averaged across several $\alpha$ levels, which rewards both coverage and sharpness (Section~\ref{section:proper_scoring_rules}; ablation in Appendix~\ref{appendix:ablations_winkler}). For \gls{fm}, we report the final checkpoint from the fixed-budget training run since early stopping was not advantageous.

See Appendix~\ref{appendix:model_archs} for full details of the two model architectures, training pipelines and hyperparameters.

\subsection{Comparison of generative and deterministic ensemble approaches}

See Appendix \ref{appendix:rollouts} (Figure~\ref{fig:rollout_crps_snapshots}) for example rollout snapshots (mean and ensemble spread).

\textbf{Overall performance.} Table \ref{tab:overall_results_table} shows that the \gls{crps}-trained ensemble is generally stronger on single-step accuracy (\gls{vrmse}, \gls{crps}) than the generative model pipeline, although the two methods perform comparably. Coverage MAE is comparable on \gls{ad} and \gls{gs}, and better for \gls{crps} on \gls{cns} and \gls{gpe}. The \gls{ssr} values also consistently favour \gls{crps} training, indicating that its predictive spread is more stable relative to its errors.
The compute trade-off goes in opposite directions for training and inference. At inference time, \gls{crps} training is substantially cheaper because each ensemble sample requires only a single forward pass (the factor difference is approximately the number of \gls{ode} steps used in \gls{fm}). By contrast, the generative model \gls{fm} trains faster per epoch because \gls{crps} training must evaluate multiple ensemble members, so the observed gap is broadly consistent with the ensemble-size overhead.

\begin{table}[t]
    \centering
    \caption{Single-step results across datasets: predictive accuracy (\acrshort{vrmse}), coverage error (MAE), probabilistic accuracy (\gls{crps}), spread--skill calibration (\gls{ssr}), inference latency, and epoch time.}
    \label{tab:overall_results_table}
\setlength{\tabcolsep}{3.5pt}
\begin{tabular}{@{}l
  l
  S[table-format=1.1e-1, detect-weight=true]
  S[table-format=1.2, detect-weight=true]
  S[table-format=1.1e-1, detect-weight=true]
  S[table-format=1.2, detect-weight=true]
  S[table-format=3.0, detect-weight=true]
  S[table-format=3.0, detect-weight=true]@{}}
\toprule
Dataset & Model & {\begin{tabular}[c]{@{}c@{}}VRMSE\\{$\downarrow$}\end{tabular}} & {\begin{tabular}[c]{@{}c@{}}Coverage\\MAE {$\downarrow$}\end{tabular}} & {\begin{tabular}[c]{@{}c@{}}CRPS\\{$\downarrow$}\end{tabular}} & {\begin{tabular}[c]{@{}c@{}}SSR\\{$\to 1$}\end{tabular}} & {\begin{tabular}[c]{@{}c@{}}Latency\\(ms/sample) {$\downarrow$}\end{tabular}} & {\begin{tabular}[c]{@{}c@{}}Epoch time\\(s) {$\downarrow$}\end{tabular}} \\
\midrule
AD & CRPS & \bfseries 1.8e-04 & 0.04 & \bfseries 8.8e-05 & \bfseries 1.07 & \bfseries 15 & 175 \\
 & FM & 2.2e-04 & \bfseries 0.02 & 1.3e-04 & 1.31 & 672 & \bfseries 24 \\
\midrule
CNS & CRPS & \bfseries 5.6e-03 & \bfseries 0.12 & \bfseries 2.7e-03 & \bfseries 1.16 & \bfseries 15 & 178 \\
 & FM & 8.7e-03 & 0.28 & 3.9e-03 & 0.65 & 715 & \bfseries 24 \\
\midrule
GS & CRPS & \bfseries 1.1e-03 & 0.20 & 5.2e-04 & \bfseries 0.67 & \bfseries 14 & 213 \\
 & FM & 1.3e-03 & \bfseries 0.18 & \bfseries 5.1e-04 & 0.57 & 687 & \bfseries 29 \\
\midrule
GPE & CRPS & \bfseries 2.8e-02 & \bfseries 0.04 & \bfseries 9.7e-03 & \bfseries 0.92 & \bfseries 15 & 181 \\
 & FM & 3.2e-02 & 0.24 & 1.1e-02 & 0.44 & 647 & \bfseries 24 \\
\bottomrule
\end{tabular}

\end{table}

See Appendix \ref{appendix:other_rollout_metrics} for performance on the autoregressive rollout task, showing the two training approaches remain broadly comparable in predictive accuracy, consistent with the overall metrics.

\begin{figure}[t]
    \centering
    \includegraphics[width=\linewidth]{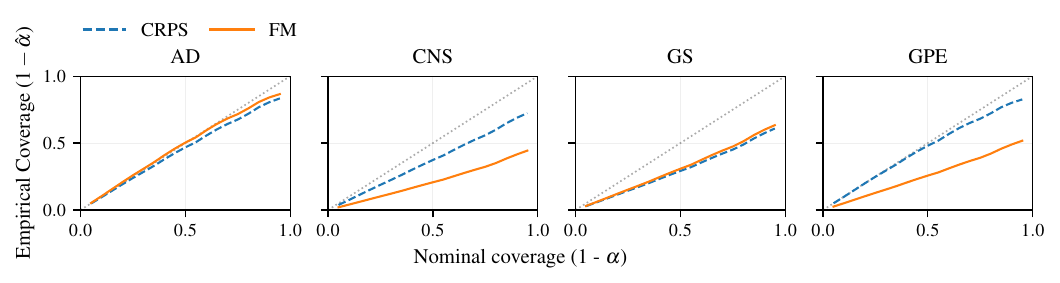}
    \caption{Coverage on the single-step prediction task by dataset (AD, CNS, GS, GPE) and model (CRPS, FM). Ideal coverage is denoted by a diagonal grey dotted line. The \gls{crps}-trained ensembles perform better than, or on par with, the generative models.}
    \label{fig:coverage_calibration_panel_overall}
\end{figure}

\textbf{UQ reliability on single-step prediction.} We first assess calibration, i.e.\ probabilistic reliability, by coverage testing for the single-step prediction task. For each model and dataset, we sweep nominal coverage levels $(1-\alpha)$ from $0.05$ to $0.95$ and plot empirical coverage $1-\hat{\alpha}$ against the nominal coverage $1-\alpha$. Perfectly calibrated forecasts would lie on the diagonal. Figure~\ref{fig:coverage_calibration_panel_overall} shows strong single-step calibration on \gls{ad} for both approaches, with curves close to the ideal diagonal. The difference between the methods is more visible on the harder datasets. On \gls{gpe}, \gls{crps} training remains reasonably well calibrated whereas the generative model clearly underestimates uncertainty. On \gls{cns}, both methods under-cover, but \gls{crps} stays noticeably closer to the target calibration line. \gls{gs} is difficult for both model classes, with both similarly under-covering.

\begin{figure*}[t]
    \centering
    \includegraphics[width=\linewidth]{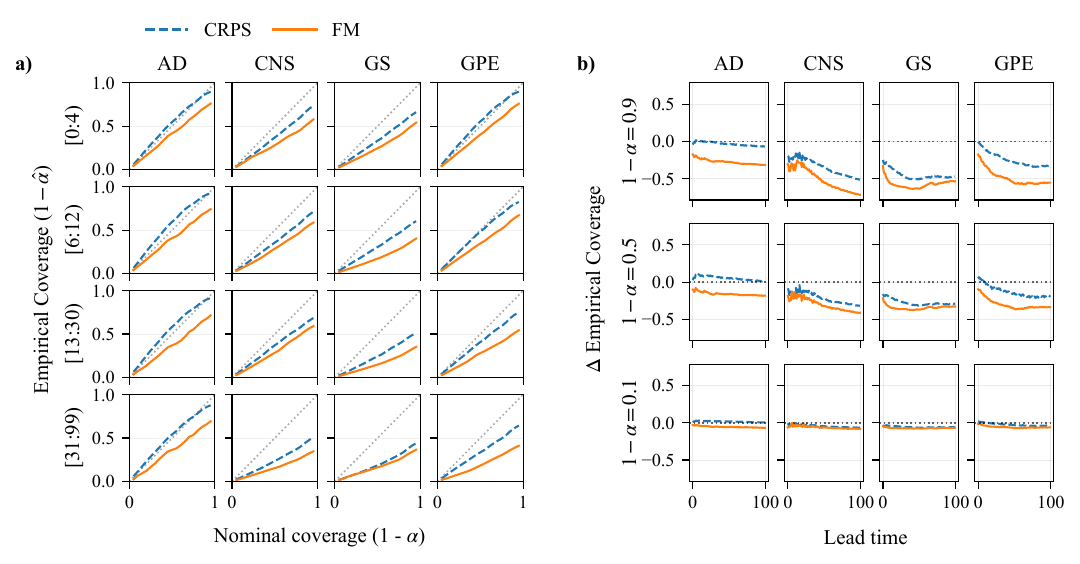}
    \caption{\gls{uq} reliability over lead time by dataset (AD, CNS, GS, GPE) and model (CRPS, FM). (a) Coverage for different lead time windows (averaging across rollout windows: $[0{:}4)$, $[6{:}12)$, $[13{:}30)$, and $[31{:}99)$). (b) $\Delta$ empirical coverage (observed minus nominal) by lead time for $1-\alpha=0.9$ (top row), $1-\alpha=0.5$ (middle row) and $1-\alpha=0.1$ (bottom row). Ideal coverage, given by zero difference, is denoted by a horizontal grey dotted line. Overall, \gls{crps} training is more reliable than the generative model approach across lead times.}
    \label{fig:uq_reliability_by_lead_time_panel}
\end{figure*}

\textbf{\gls{uq} reliability by lead time.} We next turn to \gls{uq} reliability over 100-step autoregressive rollouts. This task repeatedly feeds predictions back into the model and measures how calibration evolves over longer horizons. We report coverage averaged within rollout windows $[0{:}4)$, $[6{:}12)$, $[13{:}30)$, and $[31{:}99)$ and also show $\Delta$ empirical coverage, the signed coverage error $(1-\hat{\alpha})-(1-\alpha)$, as a function of lead time. Negative values indicate under-coverage and positive values indicate over-coverage. The $\Delta$ empirical coverage panels share a common vertical scale within each figure across nominal levels and datasets.
Figure \ref{fig:uq_reliability_by_lead_time_panel} shows that calibration typically degrades with lead time for both model classes. The main qualitative pattern from the single-step results persists, but the advantage of \gls{crps} training over the generative model approach becomes more consistent across datasets when we consider long horizon forecasts. See Appendix~\ref{appendix:test_trajectory_se} for standard errors computed over independent test trajectories.

\subsection{Ablations and other experiments}

Below we highlight the main findings from our ablations and additional experiments with further detail reported in Appendix \ref{appendix:ablations}.

\textbf{Ambient vs.\ latent space training}: \gls{crps} training in latent space does not worsen coverage, while generative training in ambient space exhibits comparable coverage to \gls{crps} training (Appendix~\ref{appendix:ablation_ambient_vs_latent}). Increasing autoencoder compression degrades generative training performance (Appendix~\ref{appendix:ablations_compression}).

\textbf{CRPS}: Using the Winkler score as a checkpoint selection criterion leads to comparable or better coverage than using the validation loss (Appendix~\ref{appendix:ablations_winkler}). Performance remained consistent when using \gls{fcrps} loss (Appendix~\ref{appendix:ablations_crpss_loss}), varying ensemble size (Appendix~\ref{appendix:ablation_crps_ensemble_size}) or reducing the number of noise channels (Appendix~\ref{appendix:ablation_crps_noise_channels}). Using backbone modulation instead of input channel concatenation to inject global simulation constants into the model worsens coverage over lead time (Appendix~\ref{appendix:ablation_crps_global_conditioning}).

\textbf{Flow matching}: Fewer \gls{ode} steps speed up inference with little change in VRMSE but can reduce coverage (Appendix~\ref{appendix:ablations_fm_ode_steps}). Exponential moving average (EMA) weights do not consistently improve coverage or rollout accuracy (Appendix~\ref{appendix:ablations_fm_ema}).

\textbf{Alternative processor architectures}: Diffusion exhibits comparable coverage to \gls{fm} while \gls{crps}-trained ensembles with a U-Net backbone exhibit worse rollout coverage on \gls{cns} and \gls{gs} relative to \gls{vit} (Appendix~\ref{appendix:ablation_architectures}). FNOs give overall less consistent rollout performance (Appendix~\ref{appendix:fno_backbone}).

\textbf{Training variability}: Retraining with an independent random seed (Appendix~\ref{appendix:training_seed}) or newly generated training data (Appendix~\ref{appendix:training_data}) shows some performance variability, most prominently for coverage on AD, but the relative performance trends and overall conclusions remain unchanged.

\textbf{MC dropout}: Using \gls{mc} dropout instead of \gls{cln} within \gls{crps} training yields comparable or worse performance (Appendix~\ref{appendix:mc_dropout_crps}). \gls{mc} dropout trained with $L_2$-regularised MSE loss is less competitive than both main approaches (Appendix~\ref{appendix:mc_dropout_mse}).

\textbf{Conformal prediction}: Since both approaches still retained miscalibrated UQ, we show in Appendix~\ref{appendix:calibration} that calibration can be improved \emph{post-hoc} using conformal prediction.

\section{Conclusions}\label{sec:conclusions}

\gls{crps}-trained ensembles generally deliver more reliable uncertainty estimates than latent space generative models. While ambient space generative modelling is often too expensive to be practical for high-dimensional problems, we found it can recover coverage comparable to \gls{crps}-trained ensembles without loss of accuracy. In contrast, \gls{crps}-trained ensembles had reliable coverage when trained in both ambient or latent space.

Our findings suggest interesting directions for future research. Since generative models are commonly trained in latent space, a natural research direction is to develop methods to improve the latent representation itself so that it preserves calibration as well as reconstruction accuracy. In particular, the current autoencoder is trained to reconstruct individual spatial fields, whereas a distributional objective would treat the decoder as a pushforward of the full conditional forecast distribution, $[d_\varphi]_\# p(\bm{z}_\mathrm{out}\mid \bm{z}_\mathrm{in}, \bm{c}) = p(\bm{x}_\mathrm{out}\mid \bm{x}_\mathrm{in}, \bm{c})$, and incorporate that distributional target directly into autoencoder training. We intend to explore this in future work.

Our study is limited to modest resolution and to a small but diverse set of physical problems due to computational constraints (a 10,000 GH200 GPU-hour allocation). While we expect the main conclusions to persist at high resolution and across more diverse problems, additional experiments are required to probe this further.

To facilitate further experimentation, we release two software packages. \texttt{AutoSim} provides a lightweight data-generation workflow for controlled \gls{pde} experiments, making it easy to generate datasets at different resolutions and simulator parameters. \texttt{AutoCast} provides the corresponding modelling framework, with support for both generative and deterministic processors, ambient and latent space training, and ensemble training with \gls{crps}-family objectives. \texttt{AutoCast} is also interoperable with \texttt{The Well}~\citep{thewell}, so models prototyped on smaller controlled simulations can be carried over to larger, diverse benchmark data with the same training pipeline.

\ifdefined\autocastanonymous\else
  \section*{Acknowledgements}

  This work was supported by the UKRI [grant number UKRI2702]. The authors acknowledge the use of resources provided by the Isambard-AI National AI Research Resource (AIRR). Isambard-AI is operated by the University of Bristol and is funded by the UK Government's Department for Science, Innovation and Technology (DSIT) via UK Research and Innovation; and the Science and Technology Facilities Council [ST/AIRR/I-A-I/1023].
\fi

\ifdefined\autocastpreprint
  \section*{AI use statement}
\else
  \subsection*{AI use statement}
\fi

AI tools assisted with the implementation of the \texttt{AutoSim}, \texttt{AutoCast} and \texttt{AutoUQ} software packages. In this way, AI contributed to research execution and synthetic data generation. We also used AI to suggest potentially relevant work based on references we had already collected, to write plotting code, to format references, and to assist with drafting and editing the manuscript. AI-assisted code was tested with unit tests and continuous integration and results were cross-checked by multiple authors. We take responsibility for this work, including text, claims or artifacts produced with the aid of generative AI.

\ifdefined\autocastpreprint\else
  \subsection*{Reproducibility statement}

  All code used in this work is released: \texttt{AutoSim} generates the four datasets, \texttt{AutoCast} implements the models, training and evaluation, and \texttt{AutoUQ} (Appendix~\ref{appendix:calibration}) implements the conformal prediction for calibration. Appendix~\ref{appendix:data_details} specifies each dataset's governing equations, numerical methods, parameter ranges and train/validation/test splits. Section~\ref{sec:experimental_setup} and Appendix~\ref{appendix:model_archs} give the architectures, hyperparameters, training budgets and checkpoint-selection criteria, and Sections~\ref{section:proper_scoring_rules} and~\ref{sec:coverage_testing} define the evaluation metrics. Appendices~\ref{appendix:test_trajectory_se}, \ref{appendix:training_seed} and~\ref{appendix:training_data} quantify variability due to the finite test set, the training seed and the training data.
\fi

\ifdefined\autocastpreprint
  \bibliographystyle{plainnat}
\else
  \bibliographystyle{iclr2027/iclr2027_conference}
\fi
\bibliography{references.bib}

\clearpage
\appendix

\section{Simulation details}\label{appendix:data_details}
\glsreset{ad}\glsreset{cns}\glsreset{gs}\glsreset{gpe}\glsreset{cln}\glsreset{ln}

All datasets were generated using the \texttt{AutoSim} package. Each uses a $64\times 64$ spatial resolution with 200 training trajectories per dataset, except \gls{gs} which has 240 training trajectories to capture all six pattern types equally, and 20 validation and 20 test trajectories (24 of each for \gls{gs}). Every trajectory has length 321 (initial state plus 320 time-steps) and is generated by randomly sampling simulation parameters from pre-defined ranges.

\subsection{Advection--Diffusion}\label{sect: adv-diff}

We consider an advection--diffusion problem describing a fluid motion in the shallow water limit
given by
\begin{subequations}\label{eq: SW_eqs}
    \begin{align}
        \frac{\partial\omega}{\partial t} + \mu \left( \frac{\partial \psi}{\partial x}\frac{\partial \omega}{\partial y} -  \frac{\partial \psi}{\partial y}\frac{\partial \omega}{\partial x}\right) & = d \nabla^2\omega, \label{eq: SW_eq_1} \\
        \nabla^2\psi                                                                                                                                                                                   & = \omega, \label{eq: SW_eq_2}           
    \end{align}
\end{subequations}
defined over a spatial domain $(x,y)\in [-L/2,L/2]^2$ and a time span $t\in [0,T]$, where $L=10$ and $T=80$.
Here $\omega(x,y,t)$ and $\psi(x,y,t)$ represent the vorticity and stream function, respectively, and $\nabla^2 = \partial_x^2+\partial_y^2$ is the two-dimensional Laplacian.
The parameter $d$ is the diffusion (viscosity-like) coefficient governing the dissipative smoothing of vorticity, while $\mu$ is a dimensionless coefficient that scales the nonlinear self-advection.

We assume periodic boundary conditions and a stretched Gaussian function as the initial condition of vorticity:
\begin{equation}
    \omega(x,y,0) = \exp\left(-x^2-\frac{y^2}{20}\right), \qquad (x,y)\in [-L/2,L/2]^2\,.
    \label{eq: w0}
\end{equation}

We are interested in approximating the time-dependent vorticity field $\omega$ as the parameters $(d,\mu)$ vary over the ranges in Table~\ref{tab:ad_params}. The vorticity equation is integrated with a finite-difference discretisation of the Laplacian and advection terms, while the Poisson equation~\eqref{eq: SW_eq_2} for the stream function is solved spectrally via \gls{fft}; time stepping is performed with an adaptive RK45 solver and a snapshot stride $\Delta t = 0.25$.

\begin{table}[h]
    \centering
    \caption{Parameter ranges used to sample \gls{ad} trajectories. Each parameter is drawn independently from a uniform distribution over the given interval.}
    \begin{tabular}{lcr}
        \toprule
        Parameter             & Symbol & Range                    \\
        \midrule
        Diffusion coefficient & $d$    & $[10^{-4}, 10^{-2}]$     \\
        Advection strength    & $\mu$  & $[0.5, 2.0]$             \\
        \bottomrule
    \end{tabular}
    \label{tab:ad_params}
\end{table}

\subsection{Conditioned Navier--Stokes}\label{sect: cns}

We consider an incompressible, buoyancy-driven \enquote{smoke-flow} system in two dimensions, in which a passive scalar $s(\bm{r}, t)$ is advected by a velocity field $\bm{u}(\bm{r}, t) = (u, v)$ and exerts a vertical buoyancy force whose strength is controlled by a conditioning parameter $\beta$. The governing equations are
\begin{subequations}\label{eq: CNS}
    \begin{align}
        \frac{\partial s}{\partial t} + \bm{u}\cdot\nabla s            & = \kappa\,\nabla^{2} s, \label{eq: CNS_smoke}                                     \\
        \frac{\partial \bm{u}}{\partial t} + (\bm{u}\cdot\nabla)\bm{u} & = -\nabla p + \nu\,\nabla^{2}\bm{u} + \big(0,\;\beta\,s\big), \label{eq: CNS_mom} \\
        \nabla\cdot\bm{u}                                              & = 0, \label{eq: CNS_div}
    \end{align}
\end{subequations}
where $\nu$ is the kinematic viscosity, $\kappa$ the smoke diffusivity, and $p$ the pressure. The buoyancy coefficient $\beta$ (named \texttt{buoyancy\_y} in the simulator) is the conditioning variable: each trajectory is generated at a fixed $\beta$ drawn from the range below, mirroring the conditioned external-force setup of PDEArena \citep{pdearena}, in which the velocity field is driven by a body force $(0, f)^{\top}$ with $f$ varied across trajectories.

The initial smoke field is a smoothed Gaussian random field constructed in Fourier space following the \texttt{phiflow} \texttt{Noise} class used by PDEArena: independent complex Gaussian coefficients are scaled by $|\bm{k}|^{-2 \sigma}$ with a low-frequency cut-off (where $\sigma$ is the \texttt{smoothness} exponent), transformed back to real space, standardised, and finally rectified ($s_0 \mapsto |s_0|$) so that the initial scalar is non-negative. The velocity is initialised to zero and is spun up purely by the buoyancy forcing acting on the smoke field.

The simulation is performed on a uniform $64\times 64$ grid over $[0, L]^{2}$ with $L = 32$ (so $\Delta x = 0.5$), using zero-gradient (Neumann) boundary conditions for the smoke and no-slip walls for the velocity, in a bounded-domain setup inspired by PDEArena (which uses $128\times 128$ at the same $L$). Time integration uses an operator-splitting scheme: (i)~semi-Lagrangian advection of $s$, $u$, and $v$ via bilinear back-tracing along the local velocity, (ii)~an explicit diffusion update for smoke and viscosity, with a vertical buoyancy term $\beta\, s$ added to the $v$-equation, and (iii)~a spectral Hodge projection onto the divergence-free subspace, which simultaneously yields the pressure $p$; in the bounded mode the projected velocity has its mean drift removed and is then explicitly zeroed at the four boundary rows/columns to enforce no-slip. The step size is adaptively limited by an advective CFL condition with $\mathrm{CFL} = 0.35$ and the explicit-diffusion stability bound. We record roughly one snapshot per integration step over a total horizon $T \approx 85.87$, yielding $n_t = 321$ frames after discarding the first $\mathrm{skip}_{n_t} = 8$ so that buoyancy-driven structure has emerged before the rollout horizon. This corresponds to PDEArena's $84$-time-unit window $[18, 102]$ but at $\sim\!6\times$ the temporal resolution (PDEArena uses $\Delta t = 1.5$, $56$ frames). The kinematic viscosity is fixed at $\nu = 0.01$ and the smoke diffusivity at $\kappa = 0$, isolating the influence of $\beta$ and the random initial condition on the resulting flow. Simulator output channels are $(s, u, v)$.

Each trajectory is parameterised by an independent uniform draw from the ranges in Table~\ref{tab:cns_params}, where $\sigma$ (\texttt{smoothness}) controls the spectral decay of the initial smoke field and $\eta$ (\texttt{noise\_scale}) sets its spectral scale.

\begin{table}[h]
    \centering
    \caption{Parameter values and ranges used to sample \gls{cns} trajectories. Ranges are sampled independently from a uniform distribution; fixed values are shown as scalars.}
    \begin{tabular}{lcr}
        \toprule
        Parameter                 & Symbol   & Value/range   \\
        \midrule
        Buoyancy coefficient      & $\beta$  & $[0.2, 0.8]$  \\
        Initial-field smoothness  & $\sigma$ & $6.0$         \\
        Initial-field noise scale & $\eta$   & $[8.0, 18.0]$ \\
        \bottomrule
    \end{tabular}
    \label{tab:cns_params}
\end{table}

\subsection{Gray--Scott}

The \gls{gs} equations are a set of coupled reaction--diffusion equations describing two chemical species, $u$ and $v$, whose scalar concentrations vary in space and time:

\begin{subequations}\label{eq: GS}
    \begin{align}
        \frac{\partial u}{\partial t} = D_u \nabla^2 u - u v^2 + F(1-u), \\
        \frac{\partial v}{\partial t} = D_v \nabla^2 v + u v^2 - (F+k)v.
    \end{align}
\end{subequations}

The parameter $F$ controls the rate at which species $u$ is added to the system and $k$ controls the rate at which species $v$ is removed (referred to as the ``feed'' and ``kill'' rates). Qualitatively different patterns emerge in the solutions depending on the two parameters $F$ and $k$ (see Table \ref{tab:gs_pattern_params} for the $F$ and $k$ values used in our simulations for each of the six pattern types. These are the same as those used in \texttt{The Well}, \citealp{thewell}).

\begin{table}[h]
    \centering
    \caption{Feed and kill parameters used to simulate each pattern type.}
    \begin{tabular}{lrr}
        \toprule
                & $F$   & $k$   \\
        \midrule
        Gliders & 0.014 & 0.054 \\
        Bubbles & 0.098 & 0.057 \\
        Maze    & 0.029 & 0.057 \\
        Worms   & 0.058 & 0.065 \\
        Spirals & 0.018 & 0.051 \\
        Spots   & 0.030 & 0.062 \\
        \bottomrule
    \end{tabular}

    \label{tab:gs_pattern_params}
\end{table}

The two constants $D_u$ and $D_v$ govern the rate of diffusion of each species. All simulations used diffusion coefficients $D_u = 2\times 10^{-5}$ and $D_v = 1\times 10^{-5}$.

The simulation domain is the square $[-L/2, L/2]^2$ with $L=1$, discretised on a uniform $64\times 64$ grid with periodic boundary conditions. The initial condition uses independent sums of randomly placed and weighted Gaussian bumps for the two species (with periodic image copies for consistency with the boundary), each normalised to $[0,1]$, with $u_0 = 1 - \mathrm{bumps}_u$ and $v_0 = \mathrm{bumps}_v$. The system is integrated with a pseudospectral exponential time-differencing Runge--Kutta scheme of order four (ETDRK4), using \glspl{fft} for the linear diffusion operator and 2/3-rule dealiasing for the nonlinear terms. We used a simulation time step $\Delta t = 1.0$ over a horizon $T = 1280$ and recorded a snapshot every 4 steps, yielding 321 frames per trajectory. Trajectories are stratified equally across the six pattern types (40 train, 4 validation, and 4 test trajectories per pattern). Some patterns occasionally produce trajectories in which a channel becomes spatially uniform at one or more frames, which destabilises normalisation-based metrics during training; we therefore reject any trajectory whose per-timestep spatial standard deviation falls below $10^{-2}$ in either channel at any frame and resample with a fresh initial condition until this criterion is met.

\subsection{Gross--Pitaevskii Equation}

The \gls{gpe} is the mean-field equation for a dilute-gas Bose--Einstein condensate (BEC), describing the dynamics of the macroscopic complex wavefunction $\psi(\bm{r}, t)$ of ultracold bosons in an external potential. We consider the two-dimensional \gls{gpe} with a moving obstacle,
\begin{equation}\label{eq: GPE}
    i\,\frac{\partial \psi}{\partial t} = \left[-\frac{1}{2}\nabla^{2} + V_{\mathrm{box}}(\bm{r}) + V_{\mathrm{obs}}(\bm{r}, t) + g\,|\psi|^{2}\right]\psi,
\end{equation}
in dimensionless harmonic-oscillator units, where $g$ controls the strength of the contact (mean-field) interaction. The condensate is confined in a flat-bottomed Woods--Saxon box trap,
\begin{equation}\label{eq: GPE_box}
    V_{\mathrm{box}}(\bm{r}) = \frac{V_{0}}{1+\exp\!\big(-(|\bm{r}| - R_{w})/a\big)},
\end{equation}
with wall height $V_{0} = 200$, wall steepness $a = 0.15$, and effective radius $R_{w} = (\mathrm{box\_param})^{-1/4} \approx 4.73$ (corresponding to $\mathrm{box\_param}=0.002$). A focused laser beam, modelled as a Gaussian repulsive obstacle, sweeps sinusoidally back and forth along the $y$-axis,
\begin{equation}\label{eq: GPE_obs}
    V_{\mathrm{obs}}(\bm{r}, t) = U\exp\!\left(-\frac{x^{2} + (y - R\sin(\omega t))^{2}}{w^{2}}\right),
\end{equation}
with peak speed $v_{\max} = R\,\omega$. Above a critical obstacle velocity $v_c$ the superfluid response transitions from dissipationless laminar flow \citep{raman1999} to periodic vortex--antivortex pair shedding (a quantum von K\'arm\'an street; \citealp{sasaki2010,kwon2016}) and on to increasingly disordered wakes that follow a superfluid-Reynolds-number scaling \citep{reeves2015}. The parameter ranges below are chosen so that $v_{\max}$ straddles $v_c$. Because the obstacle motion is sinusoidal, a single sweep can pass through slower and faster phases of the forcing.

The simulation is performed on a uniform $64\times 64$ periodic Fourier grid over the spatial domain $[-L/2, L/2]^{2}$ with $L=10$, with physical confinement provided by the Woods--Saxon box potential. We use a Strang split-step Fourier scheme \citep{bao2006} that alternates combined potential and nonlinear half-steps in real space with a kinetic full-step in Fourier space with 2/3-rule dealiasing. Each trajectory consists of two phases: (i) imaginary-time relaxation for $4000$ steps at $\Delta t_{\mathrm{imag}} = 0.0025$ from a broad Gaussian seed centred at the origin, which produces a clean vortex-free ground state inside the box; followed by (ii) real-time evolution for $T = 17.55$ time units at $\Delta t = 0.005$, with snapshots recorded every $\Delta t_{\mathrm{snap}} = 0.05$ and the first $30$ frames discarded so that the obstacle is well into its first sweep. Simulator output channels are the $\Re(\psi)$ and $\Im(\psi)$ fields, enabling both density and phase to be captured by the emulator indirectly. $\Re(\psi)$ and $\Im(\psi)$ are jointly normalised so that phase is unaffected.

Each trajectory is parameterised by an independent uniform draw from the ranges in Table~\ref{tab:gpe_params}, chosen so that the resulting trajectories span a range of dynamical complexity, from largely laminar response with linear density excitations, through periodic vortex--antivortex pair shedding, to disordered wakes populated by many vortex cores.

\begin{table}[h]
    \centering
    \caption{Parameter ranges used to sample \gls{gpe} trajectories. Each parameter is drawn independently from a uniform distribution over the given interval.}
    \begin{tabular}{lcr}
        \toprule
        Parameter                      & Symbol   & Range        \\
        \midrule
        Interaction strength           & $g$      & $[80, 200]$  \\
        Obstacle strength              & $U$      & $[3, 8]$     \\
        Obstacle angular frequency     & $\omega$ & $[0.5, 2.0]$ \\
        Obstacle Gaussian $1/e$ radius & $w$      & $[0.8, 2.5]$ \\
        Obstacle sweep amplitude       & $R$      & $[2.0, 3.0]$ \\
        Initial Gaussian width         & $w_{0}$  & $[1.5, 2.0]$ \\
        \bottomrule
    \end{tabular}
    \label{tab:gpe_params}
\end{table}

\section{Model architectures and hyperparameters}\label{appendix:model_archs}

In this section we provide further detail on training and model architectures.

\textbf{Training}

We keep the \gls{crps} and \gls{fm} training as aligned as possible. The nominal batch size is fixed at $1024$. For \gls{crps} training, the input batch is replicated across ensemble members, so the effective number of distinct training examples scales as $1024/M$ for ensemble size $M$; for example, with $M=4$, the effective batch size is $256$. We use a cosine learning-rate schedule decaying to zero over the precomputed number of epochs allowed by the time budget, with learning rate $2\times 10^{-4}$ for \gls{crps} training and $10^{-4}$ for \gls{fm}.

\textbf{Autoencoder}

For latent space experiments, we follow the deep compression autoencoder design used in \citet{lola}. Each dataset has its own $49$M parameter autoencoder trained for $512$ epochs with learning rate $10^{-5}$, a cosine schedule, and a variance-scaled mean-squared reconstruction loss. The encoder has three resolution levels with three residual blocks per level, $3\times3$ convolutions, \gls{ln}, and dropout $0.05$. Hidden widths follow the standard convention of doubling at each downsample, giving $[128, 256, 512]$ channels at successive levels, which keeps per-level compute balanced and provides more channel capacity at coarser spatial scales. Stride-$2$ pixel-shuffle downsampling is applied between consecutive levels, so the two downsamples (between levels $1$--$2$ and $2$--$3$) reduce the spatial resolution by a factor of $4$ and take the input from $64\times64$ at level $1$ to $16\times16$ at level $3$. The decoder mirrors this with widths $[512, 256, 128]$ and pixel-shuffle upsampling; the latent bottleneck is softly clipped with scale $5$ to avoid unbounded latent amplitudes. Overall the encoder compresses states from $64\times64\times C_{\text{in}}$ to $16\times16\times C_{\text{latent}}$ with $C_{\text{latent}}=8$. This corresponds to a relatively mild compression rate compared with some recent latent-emulation settings, but is appropriate for our $64\times64$ resolution regime and allows us to isolate the effect of latent space forecasting without introducing an aggressively lossy bottleneck.

\textbf{Processors}

Our main comparison is between two model classes with matched \gls{vit} backbones. The first is a stochastic generative model trained in latent space with a \gls{fm} objective~\citep{flowmatching}. The second is a deterministic backbone trained probabilistically using noise injection and \gls{afcrps} loss~\citep{aifscrps} in ambient space. By default, the \gls{crps} model uses $M=8$ ensemble members, with additional ensemble sizes considered in the ablations. For the \gls{crps} ensemble, stochasticity is introduced through \gls{cln} with adaLN-Zero-style modulation. Because the injected noise channels add parameters, we reduce the hidden width of the \gls{crps} processor slightly so that its size remains comparable to the flow-matching processor. \gls{fm} sampling uses Euler integration; the step count is ablated in Appendix~\ref{appendix:ablations_fm_ode_steps}.

Table~\ref{tab:model_architecture_summary} summarises the processor architectures used in the main comparison and most of the ablations, including with U-Net and diffusion models. The FNO and \gls{mc}-dropout variants are specified in their dedicated appendices (Appendices~\ref{appendix:fno_backbone}--\ref{appendix:mc_dropout_mse}). 

The \gls{vit} processors share the same backbone hyperparameters ($12$ transformer blocks, $8$ attention heads) and operate on a $16\times16$ token grid in both spaces, using patch size $4$ on $64\times64$ ambient states and patch size $1$ on $16\times16$ autoencoder latents. The autoencoder compression ablation (Appendix~\ref{appendix:ablations_compression}) is the one exception, where an $8\times8\times8$ latent yields an $8\times8$ token grid; all other rows of Table~\ref{tab:model_architecture_summary} use the $16\times16$ grid. In both the \gls{fm}/diffusion and \gls{crps} pipelines, the single input frame and four output frames are folded into the channel dimension at the processor boundary, so the backbone always runs with a single time token; temporal mixing is therefore mediated entirely through the channel axis. The \gls{fm} and diffusion \glspl{vit} use width $704$; the \gls{crps}-trained \gls{vit} uses width $568$ to keep the parameter count near $80$M after adding the $1024$-dimensional noise modulation path. Reported counts are processor-only and exclude the autoencoder.

\begin{table}[t]
    \centering
    \caption{Processor architectures used in the main comparison and most of the ablations. The first two columns are the main comparison; the remaining four columns are architecture ablations. The right-most column is the only \gls{crps} variant that uses a U-Net backbone instead of a \gls{vit}. Dashes mark settings that do not apply. Parameter counts are for the processor and exclude the autoencoder; reported values are for \gls{cns} ($C=3$) and vary by less than $0.3$M across the $C\in\{1,2,3\}$ datasets, since state-channel count only enters the patch embedding and output head. Latent processors operate on the autoencoder latent ($C_{\text{latent}}=8$) and are identical across datasets.}
    \label{tab:model_architecture_summary}
    \small
    \setlength{\tabcolsep}{4pt}
    \begin{tabularx}{\linewidth}{@{}l*{6}{>{\centering\arraybackslash}X}@{}}
        \toprule
                           & \multicolumn{2}{c}{Main} & \multicolumn{4}{c}{Ablations} \\
        \cmidrule(lr){2-3}\cmidrule(lr){4-7}
                           & \gls{fm}    & \gls{crps}  & \gls{fm}    & \gls{crps}  & Diffusion   & \gls{crps} \\
                           & (latent)    & (ambient)   & (ambient)   & (latent)    & (latent)    & (ambient) \\
        \midrule
        Backbone           & \gls{vit}   & \gls{vit}   & \gls{vit}   & \gls{vit}   & \gls{vit}   & U-Net \\
        Width / channels   & $704$       & $568$       & $704$       & $568$       & $704$       & 62, 124, 248, 496 \\
        Blocks             & $12$        & $12$        & $12$        & $12$        & $12$        & 3, 3, 3, 3 \\
        Heads              & $8$         & $8$         & $8$         & $8$         & $8$         & --- \\
        Patch size         & $1$         & $4$         & $4$         & $1$         & $1$         & --- \\
        MLP expansion      & $4$         & $4$         & $4$         & $4$         & $4$         & $1$ \\
        Noise channels     & ---         & $1024$      & ---         & $1024$      & ---         & $1024$ \\
        Ensemble $M$       & ---         & $8$         & ---         & $8$         & ---         & $8$ \\
        Sampler / steps    & Euler / 50  & ---         & Euler / 50  & ---         & Euler / 50  & --- \\
        Loss               & \gls{fm}    & \gls{afcrps} & \gls{fm}    & \gls{afcrps} & Karras DM   & \gls{afcrps} \\
        Parameters         & $80.0$M     & $80.9$M     & $80.2$M     & $81.8$M     & $80.0$M     & $81.3$M \\
        \bottomrule
    \end{tabularx}
\end{table}

\textbf{Simulation constants and conditioning}

For each trajectory, the simulator parameters $\bm{c}\in\mathbb{R}^{C_c}$ are sampled once and held fixed for the entire rollout, including under autoregressive prediction. The state is normalised, and (for latent processors) passed through the autoencoder; $\bm{c}$ is not encoded and is used only for conditioning. We inject $\bm{c}$ into the processor in one of two ways.

\emph{Channel concatenation} (used in the main ambient \gls{crps} baseline). Each scalar in $\bm{c}$ is broadcast across the $W\times H$ grid and concatenated to the current state $\bm{x}^{i}$ along the channel axis, giving the processor input
\begin{equation}
\tilde{\bm{x}} \;=\; \bm{x}^{i} \,\|\, \bm{c} \;\in\; \mathbb{R}^{W\times H\times (C+C_c)},
\end{equation}
where $\|$ denotes concatenation along the channel axis (the spatial broadcast of $\bm{c}$ is implicit in the shape). Backbone modulation by $\bm{c}$ is then disabled so that $\bm{c}$ is not injected twice.

\emph{Backbone modulation} (used in latent \gls{fm}, latent \gls{crps}, and diffusion). The vector $\bm{c}$ is embedded by a small MLP and added to the timestep/noise modulation vector that drives the per-block modulation in the backbone (see equations below). The state itself is not channel-augmented with $\bm{c}$.

The global-conditioning ablation in Appendix~\ref{appendix:ablation_crps_global_conditioning} is the only run where the ambient \gls{crps} model is switched from channel concatenation to backbone modulation, with all other settings of Table~\ref{tab:model_architecture_summary} unchanged.

\textbf{Backbone modulation}

We use the \gls{vit} and U-Net backbones implemented in \texttt{Azula} \citep{azula}. They both use a single shared adaptive \gls{ln} per block. Letting $F$ denote the block branch (self-attention followed by an MLP in the \gls{vit}; optional spatial attention followed by a convolutional feed-forward block in the U-Net), the per-block computation mapping input $\bm{h}$ to output $\bm{h}^{+}$ is
\[
\begin{aligned}
\tilde{\bm{h}} &= (1+a(\bm{\eta}))\odot\mathrm{LN}(\bm{h})+b(\bm{\eta}),\\
\bm{h}^{+} &= \bigl(\bm{h}+g(\bm{\eta})\odot F(\tilde{\bm{h}})\bigr)\bigl(1+g(\bm{\eta})^{2}\bigr)^{-1/2},\\
[a,b,g] &= A(\bm{\eta}),
\end{aligned}
\]
with $\mathrm{LN}$ a non-affine layer normalisation, $A$ a two-layer MLP (Linear--SiLU--Linear) whose final-layer weights are multiplied by $10^{-2}$ at initialisation, and $(a,b,g)$ shared across the attention and MLP sub-layers within a block (rather than the two-modulation variant of the canonical adaLN-Zero~\citep{dit}). The variance-preserving factor $(1+g(\bm{\eta})^{2})^{-1/2}$ keeps the residual norm bounded as $g$ grows during training.

\section{Example rollouts}\label{appendix:rollouts}

\begin{figure}[H]
    \centering
    \begin{subfigure}[t]{0.49\linewidth}
        \centering
        \includegraphics[width=\linewidth]{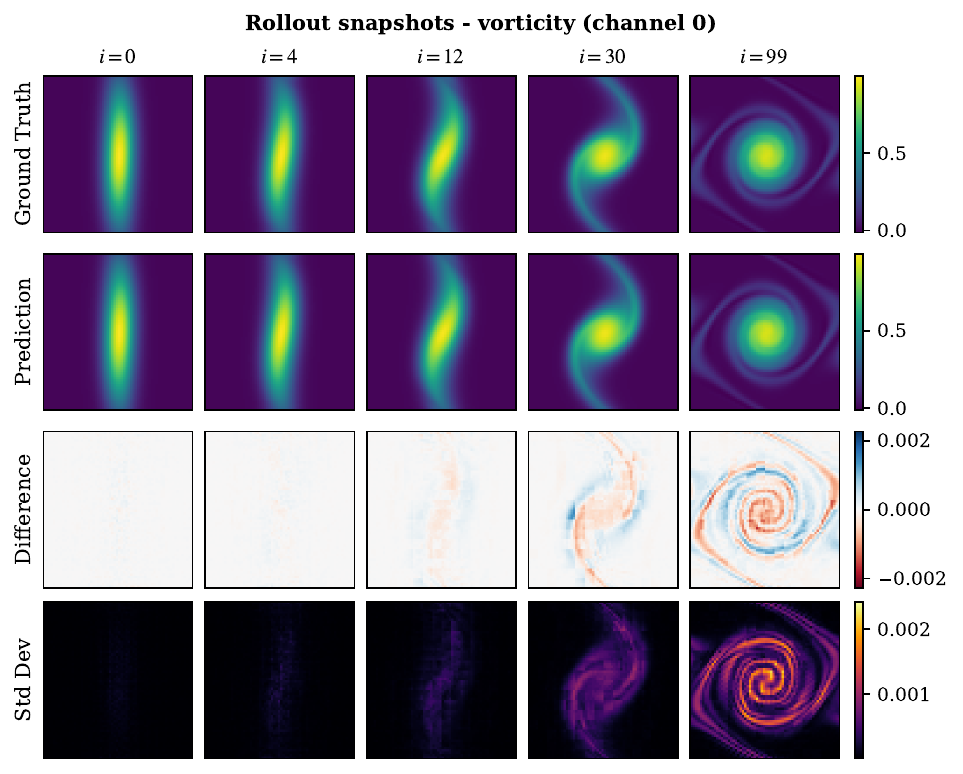}
        \caption{AD}
        \label{fig:rollout_snap_ad}
    \end{subfigure}
    \hfill
    \begin{subfigure}[t]{0.49\linewidth}
        \centering
        \includegraphics[width=\linewidth]{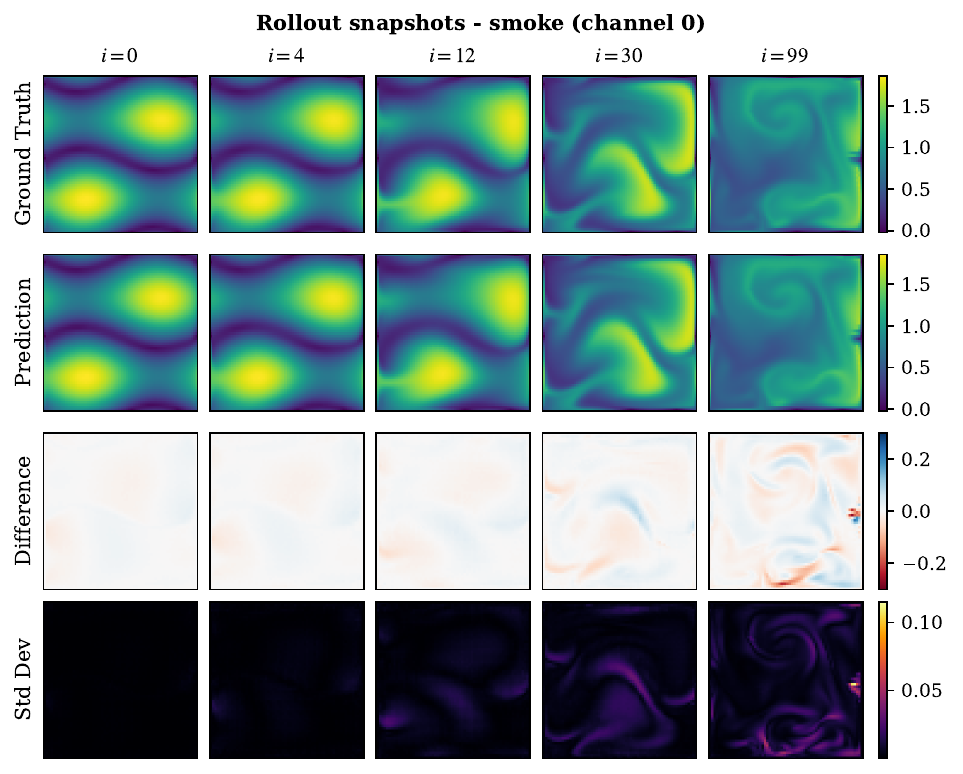}
        \caption{CNS}
        \label{fig:rollout_snap_cns}
    \end{subfigure}

    \vspace{0.75em}

    \begin{subfigure}[t]{0.49\linewidth}
        \centering
        \includegraphics[width=\linewidth]{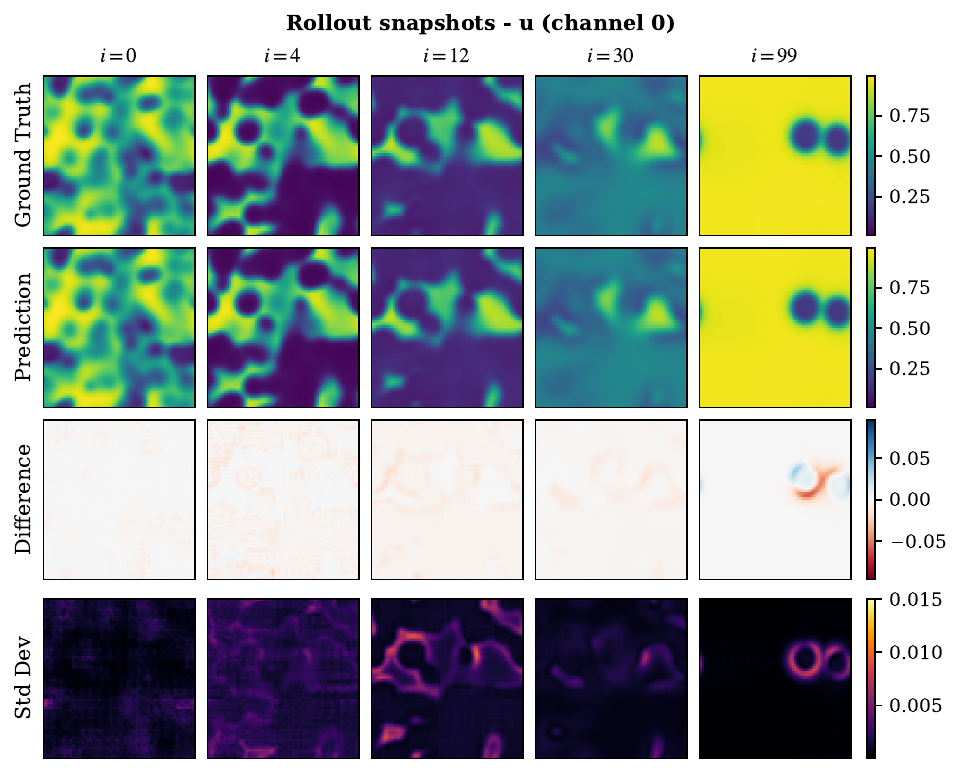}
        \caption{GS}
        \label{fig:rollout_snap_gs}
    \end{subfigure}
    \hfill
    \begin{subfigure}[t]{0.49\linewidth}
        \centering
        \includegraphics[width=\linewidth]{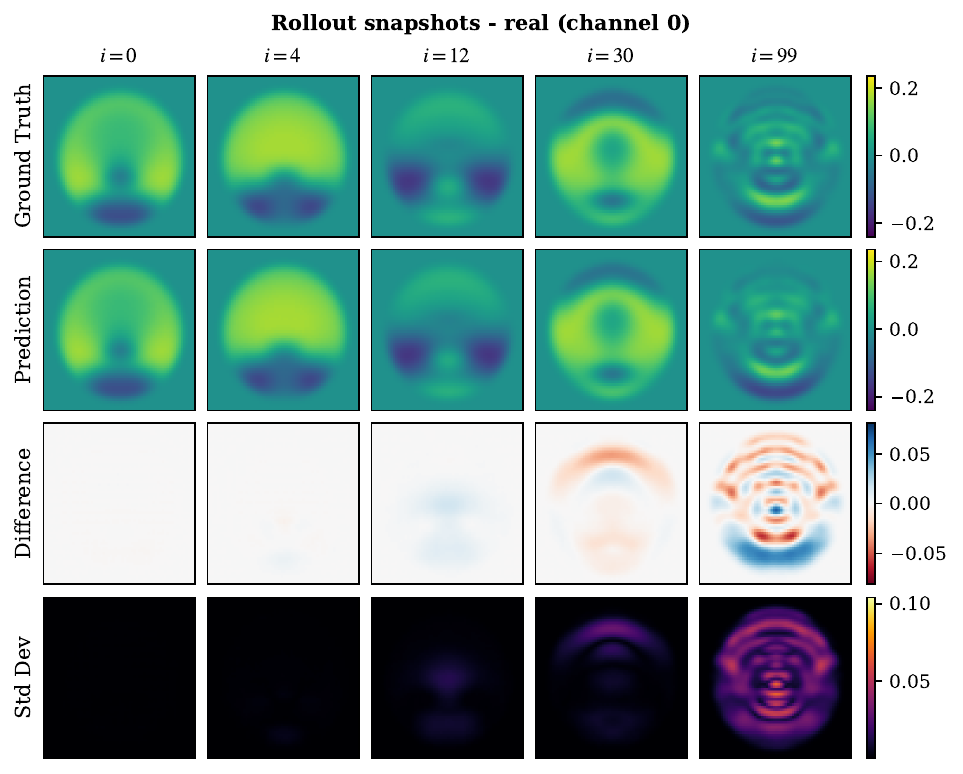}
        \caption{GPE}
        \label{fig:rollout_snap_gpe}
    \end{subfigure}
    \caption{Example rollout snapshots (channel $0$) for one test case per dataset. Top to bottom row: the ground-truth trajectory, ensemble mean, difference between ground truth and ensemble mean, and ensemble standard deviation.}
    \label{fig:rollout_crps_snapshots}
\end{figure}

\section{Other rollout metrics}\label{appendix:other_rollout_metrics}

\begin{figure}[H]
    \centering
    \includegraphics[width=\linewidth]{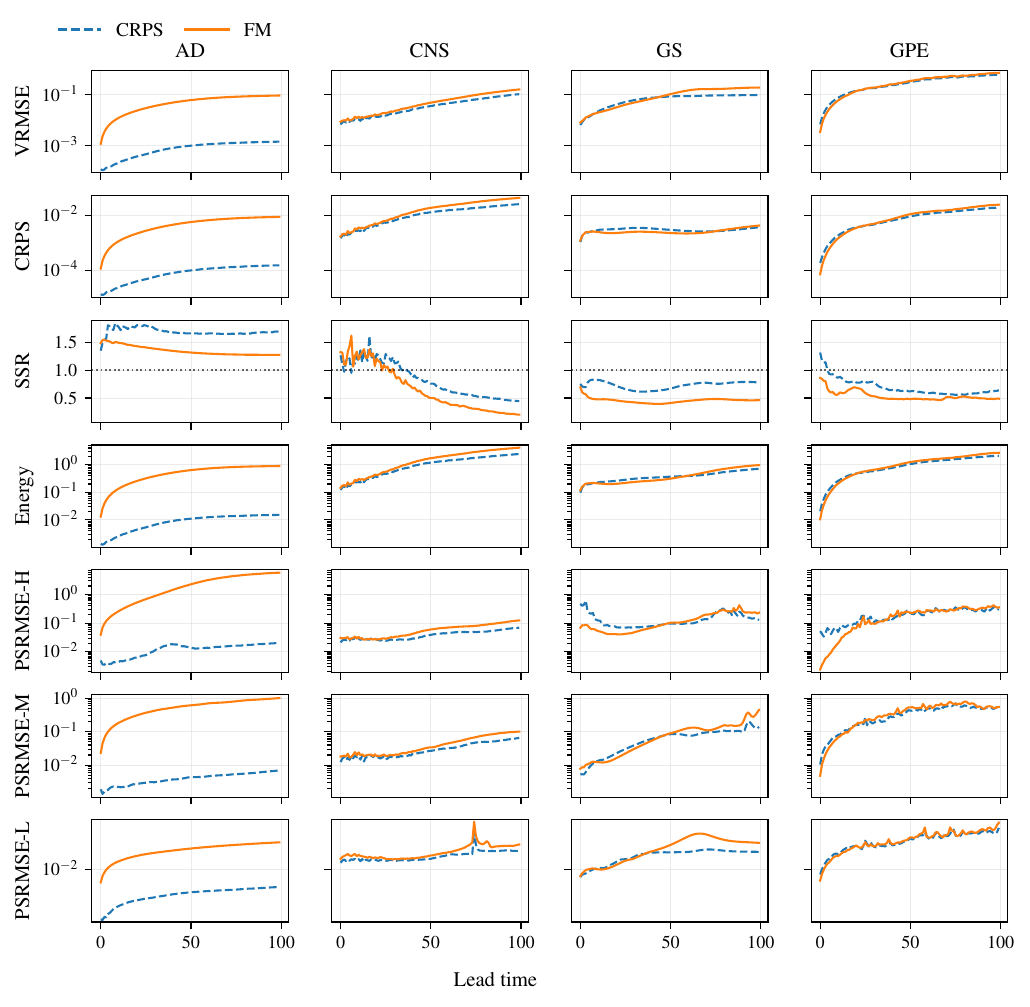}
    \caption{Top to bottom: \gls{vrmse}, \gls{crps}, \gls{ssr}, Energy score, power spectrum RMSE (high), power spectrum RMSE (medium), power spectrum RMSE (low) over lead time by dataset (\gls{ad}, \gls{cns}, \gls{gs}, \gls{gpe}) and model (CRPS, FM). The two approaches have comparable accuracy across metrics while the \gls{crps} training approach tends to have better \gls{ssr} except on the \gls{ad} dataset.}
    \label{fig:lead_time_panel_crps}
\end{figure}

Figure \ref{fig:lead_time_panel_crps} shows a number of accuracy and calibration metrics measuring performance over 100-step autoregressive rollouts. We report the same metrics as those used for single-step performance evaluation in the main text (\gls{ssr}, \gls{crps} and \gls{vrmse}) as well as the energy score (a multivariate generalisation of the \gls{crps}) and \gls{psrmse} at different spatial frequencies (following \citealp{lola}).

The results indicate that the two training approaches remain broadly comparable in predictive accuracy across a range of metrics. As expected, error increases with lead time for both methods. \gls{ad} is the clearest separation in favour of \gls{crps}, where both models remain accurate but the \gls{crps} rollout error is one to two orders of magnitude lower. On \gls{cns} and \gls{gs}, \gls{crps} also retains a modest accuracy advantage at larger lead times. The \gls{crps} training approach also tends to have better \gls{ssr} over long lead times, except on the \gls{ad} dataset.

\clearpage
\section{Finite-test-trajectory standard errors}\label{appendix:test_trajectory_se}

We regenerated predictions while retaining trajectory identifiers and quantify sampling uncertainty from the finite test set conditional on each trained model. Since windows, output times, spatial locations, and channels from the same rollout are dependent, the independently generated test trajectory is the sampling unit. For a scalar metric, let $U_i$ denote the statistic obtained after first aggregating all relevant predictions within test trajectory $i$. For $n$ trajectories, we report
\begin{equation}
    \overline{U}=\frac{1}{n}\sum_{i=1}^{n}U_i,
    \qquad
    \widehat{\mathrm{SE}}(\overline{U})
    =\sqrt{\frac{1}{n(n-1)}\sum_{i=1}^{n}(U_i-\overline{U})^2}.
\end{equation}
This calculation is used for \gls{vrmse}, \gls{crps}, \gls{ssr}, energy score, and the spectral error metrics. For \gls{gs}, which has six equally represented pattern strata, we preserve the sampling design: if stratum $h$ has weight $w_h$, size $n_h$, mean $\overline{U}_h$, and sample variance $s_h^2$, then
\begin{equation}
    \overline{U}_{\mathrm{strat}}=\sum_{h=1}^{6}w_h\overline{U}_h,
    \qquad
    \widehat{\mathrm{SE}}(\overline{U}_{\mathrm{strat}})
    =\left(\sum_{h=1}^{6}\frac{w_h^2s_h^2}{n_h}\right)^{1/2}.
\end{equation}
In our \gls{gs} test set the strata are equally weighted, so $w_h=1/6$, and each contains $n_h=4$ trajectories.

Coverage MAE requires an additional covariance calculation because coverage at different nominal levels is evaluated on the same trajectories. Let $c_{il}$ be the empirical coverage for trajectory $i$ at nominal level $q_l$, where $l\in\{1,\ldots,L\}$, so that $\bm{c}_i=(c_{i1},\ldots,c_{iL})\in\mathbb{R}^{L}$ and $\overline{c}_l=n^{-1}\sum_i c_{il}$. Define coverage MAE as the scalar function
\begin{equation}
    g(\overline{\bm{c}})
    =\frac{1}{L}\sum_{l=1}^{L}\lvert\overline{c}_l-q_l\rvert,
    \qquad
    \partial_l g(\overline{\bm{c}})
    =\frac{1}{L}\operatorname{sign}(\overline{c}_l-q_l).
\end{equation}
For the unstratified datasets, the estimated covariance of the mean coverages is
\begin{equation}
    \widehat{\operatorname{Cov}}(\overline{c}_l,\overline{c}_m)
    =\frac{1}{n(n-1)}\sum_{i=1}^{n}
        (c_{il}-\overline{c}_l)(c_{im}-\overline{c}_m).
\end{equation}
The stratified analogue is used for \gls{gs}. The delta method then gives
\begin{equation}
    \widehat{\mathrm{SE}}\!\left[g(\overline{\bm{c}})\right]
    =\left[
        \partial_l g(\overline{\bm{c}})\,
        \widehat{\operatorname{Cov}}(\overline{c}_l,\overline{c}_m)\,
        \partial_m g(\overline{\bm{c}})
    \right]^{1/2},
\end{equation}
where repeated level indices $l$ and $m$ are summed; $i$ remains the trajectory index and is summed explicitly in the covariance estimator. This retains the covariance across nominal coverage levels.

\begin{figure}[H]
    \centering
    \includegraphics[width=\linewidth]{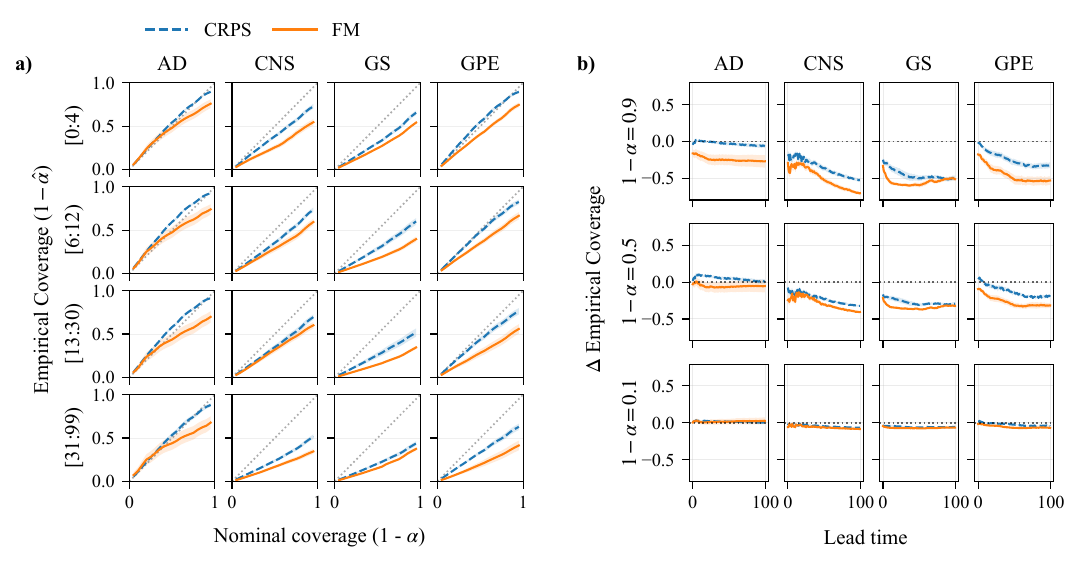}
    \caption{Coverage with finite-test-trajectory standard errors for the main comparison across datasets (\gls{ad}, \gls{cns}, \gls{gs}, \gls{gpe}). Lines show trajectory-aggregated estimates and bands show plus or minus one standard error. Panels report (a) empirical vs.\ nominal coverage over rollout windows and (b) $\Delta$ empirical coverage (observed minus nominal) at three nominal levels over rollout lead time. The \gls{gs} bands account for its six pattern strata.}
    \label{fig:main_comparison_test_trajectory_se_ablation}
\end{figure}

\begin{table}[H]
    \centering
    \caption{Single-step main-comparison results reported as mean (finite-test-trajectory standard error). Bold denotes the better result for each dataset.}
    \label{tab:main_comparison_test_trajectory_se}
    \footnotesize
\setlength{\tabcolsep}{2.5pt}
\begin{tabular}{@{}llcccc@{}}
\toprule
Dataset & Model & VRMSE $\downarrow$ & \begin{tabular}[c]{@{}c@{}}Coverage MAE\\$\downarrow$\end{tabular} & CRPS $\downarrow$ & SSR $\to 1$ \\
\midrule
AD  & CRPS & \bfseries 1.8e-04 (2.2e-05) & 0.041 (0.009) & \bfseries 8.8e-05 (1.3e-05) & \bfseries 1.07 (0.03) \\
    & FM   & 2.2e-04 (3.2e-05) & \bfseries 0.019 (0.003) & 1.3e-04 (2.0e-05) & 1.30 (0.04) \\
\midrule
CNS & CRPS & \bfseries 5.6e-03 (1.2e-04) & \bfseries 0.124 (0.007) & \bfseries 2.7e-03 (8.6e-05) & \bfseries 1.16 (0.04) \\
    & FM   & 8.7e-03 (3.5e-04) & 0.284 (0.006) & 3.9e-03 (1.5e-04) & 0.66 (0.03) \\
\midrule
GS  & CRPS & \bfseries 1.1e-03 (8.4e-05) & 0.198 (0.008) & 5.2e-04 (7.2e-05) & \bfseries 0.67 (0.01) \\
    & FM   & 1.3e-03 (5.3e-05) & \bfseries 0.183 (0.006) & \bfseries 5.1e-04 (2.3e-05) & 0.57 (0.01) \\
\midrule
GPE & CRPS & \bfseries 2.8e-02 (2.9e-03) & \bfseries 0.034 (0.010) & \bfseries 9.6e-03 (1.2e-03) & \bfseries 0.92 (0.05) \\
    & FM   & 3.2e-02 (3.4e-03) & 0.24 (0.01) & 1.1e-02 (1.6e-03) & 0.43 (0.03) \\
\bottomrule
\end{tabular}

\end{table}

The compact coverage comparison in Figure~\ref{fig:main_comparison_test_trajectory_se_ablation} and the single-step results in Table~\ref{tab:main_comparison_test_trajectory_se} leave the qualitative comparison unchanged: \gls{crps} retains substantially better coverage on \gls{cns} and \gls{gpe}, while the smaller \gls{ad} and \gls{gs} differences remain dataset dependent.

\clearpage

\begin{figure}[H]
    \centering
    \includegraphics[width=\linewidth]{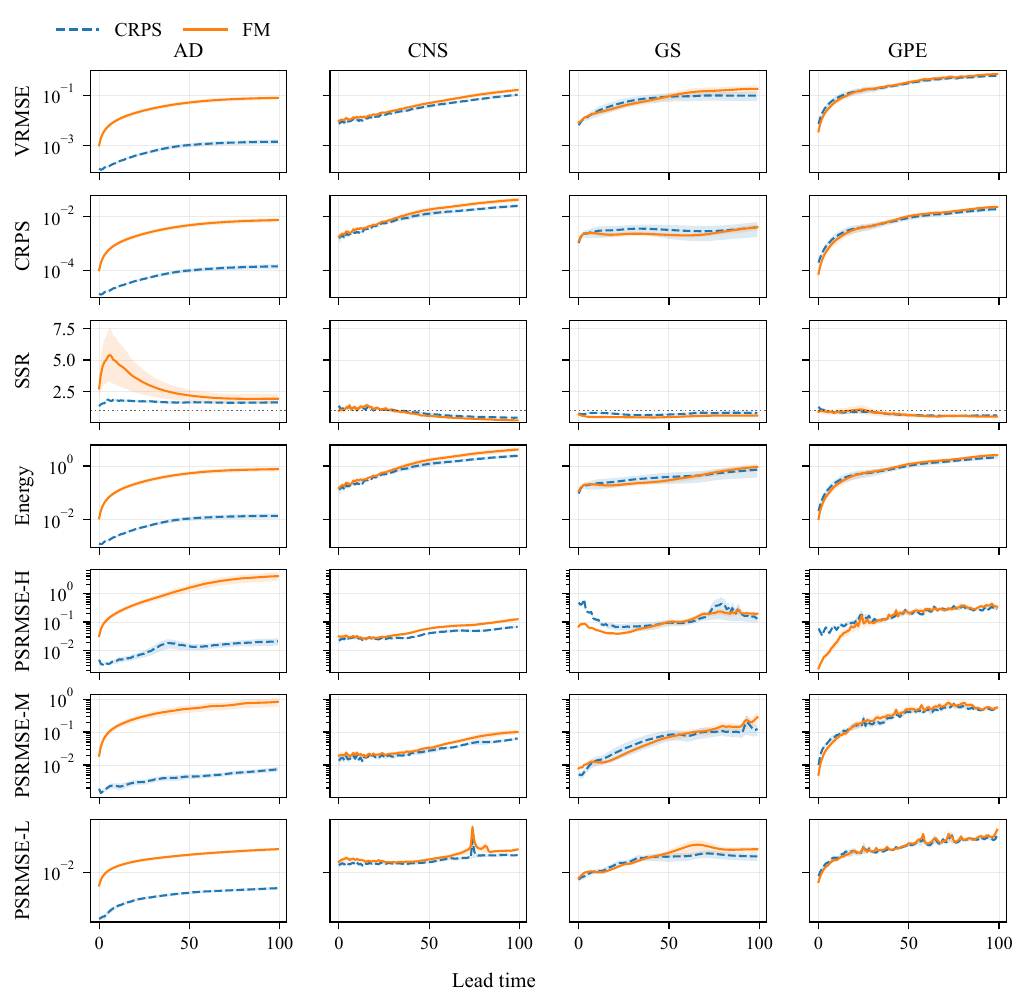}
    \caption{Finite-test-trajectory uncertainty for the main \gls{crps} and \gls{fm} comparison across datasets (\gls{ad}, \gls{cns}, \gls{gs}, \gls{gpe}). Lines show the trajectory-aggregated metric and shaded bands show plus or minus one standard error over test trajectories. Rows report \gls{vrmse}, \gls{crps}, \gls{ssr}, energy score, and high-, mid-, and low-frequency power-spectrum RMSE over rollout lead time. The \gls{gs} bands account for its six pattern strata.}
    \label{fig:main_comparison_test_trajectory_se}
\end{figure}

Figure~\ref{fig:main_comparison_test_trajectory_se} provides the corresponding standard-error bands for accuracy, probabilistic scores, and spectral errors. These standard errors quantify finite-test-trajectory sampling only and exclude training-seed and training-data variability (Appendices~\ref{appendix:training_seed} and~\ref{appendix:training_data}).

\section{Ablations and other experiments}\label{appendix:ablations}

Unless noted otherwise, each ablation below varies a single setting (named by its subsection title) from the corresponding column of Table~\ref{tab:model_architecture_summary}; all other architectural choices, training schedule, and evaluation protocol follow the main runs (Section~\ref{sec:experimental_setup}).

\subsection{Latent vs.\ ambient space training}\label{appendix:ablation_ambient_vs_latent}

We consider \gls{fm} training in ambient space and \gls{crps} training in latent space, providing counterparts to the results in the main comparison. Applying \gls{fm} directly in ambient space follows by substituting $\bm{x}_{\mathrm{in}}$, $\bm{x}_{\mathrm{out}}$, and $\bm{x}_t$ for $\bm{z}_{\mathrm{in}}$, $\bm{z}_{\mathrm{out}}$, and $\bm{z}_t$ as outlined in Section~\ref{methods:generative_models}, without the need for encoding and decoding. Applying \gls{crps} in latent space uses the same latent representation as \gls{fm}, with the \gls{crps} loss calculated directly on the latent space predictions $p_\theta(\bm{z}_\mathrm{out}|\bm{z}_\mathrm{in}, \bm{c})$.

\begin{figure}[H]
    \centering
    \includegraphics[width=\linewidth]{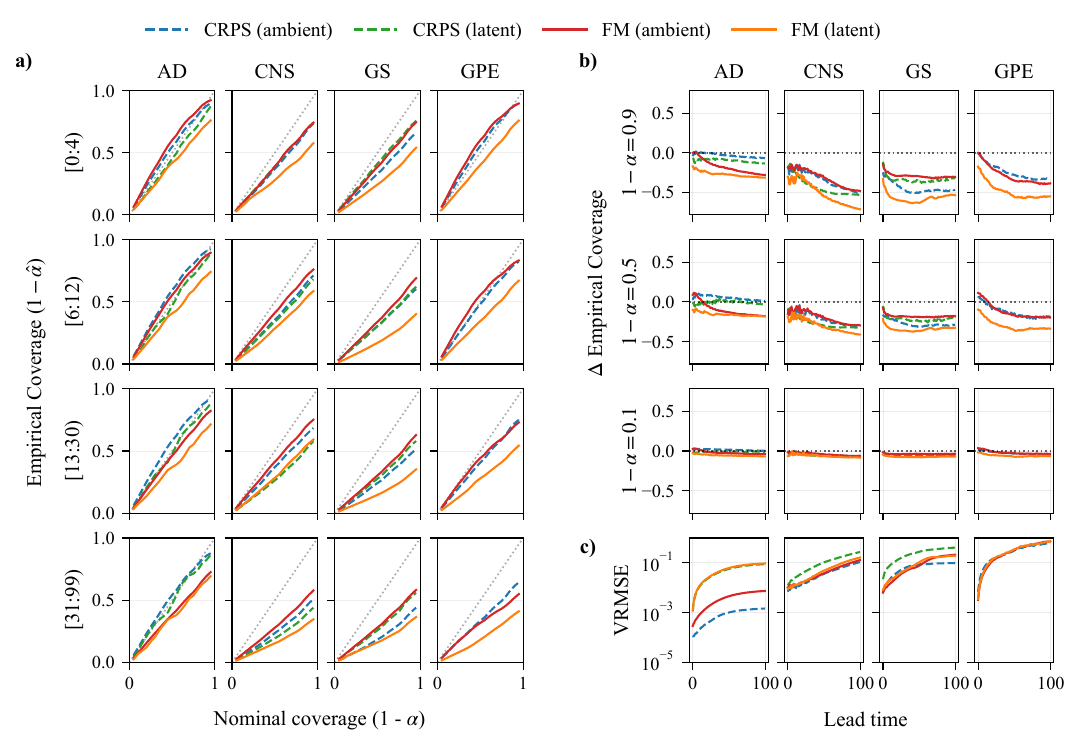}
    \caption{Latent vs.\ ambient training: \gls{fm} (ambient vs.\ latent) and \gls{crps} training (ambient vs.\ latent), $M=8$, across datasets (\gls{ad}, \gls{cns}, \gls{gs}, \gls{gpe}). (a) Empirical vs.\ nominal coverage averaged over rollout windows. (b) $\Delta$ empirical coverage (observed minus nominal) by lead time over $100$-step rollouts. (c) \gls{vrmse} by lead time. The latent \gls{fm} model has consistently worse coverage while all the other approaches perform comparably. No result is reported for the CRPS model trained in latent space on the \gls{gpe} dataset because the ensemble collapsed.}
    \label{fig:ablation_latent}
\end{figure}

Figure~\ref{fig:ablation_latent} shows that moving \gls{fm} from latent space back to ambient space substantially improves its coverage and makes it broadly comparable to \gls{crps} training. The corresponding shift for \gls{crps} is much smaller: training \gls{crps} in latent space causes only a limited degradation in coverage, although it does increase \gls{vrmse} for \gls{cns} and \gls{gs}. For \gls{gpe}, the latent space \gls{crps} run collapsed entirely, so in that case latent training is not a viable substitute for the ambient space model. All variants perform well on \gls{ad}, although the \gls{crps}-trained models are still more stable at large lead times.

\clearpage
\subsection{Increased autoencoder compression}\label{appendix:ablations_compression}

\begin{figure}[H]
    \centering
    \includegraphics[width=0.9\linewidth]{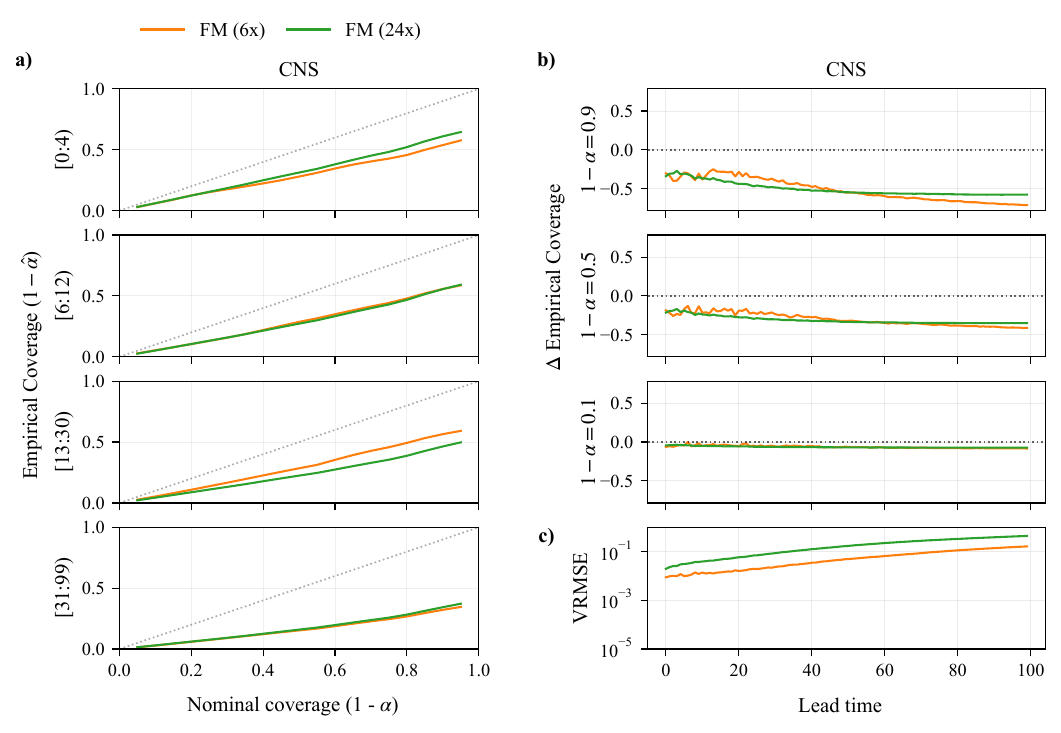}
    \caption{\gls{fm} performance on \gls{cns} at increased autoencoder compression ($8\times 8\times 8$ latent, 24x compression) vs.\ the 6x compression used in the main results. (a) Empirical vs.\ nominal coverage averaged over rollout windows. (b) $\Delta$ empirical coverage (observed minus nominal) by lead time over $100$-step rollouts. (c) \gls{vrmse} by lead time. Increasing the autoencoder compression rate further degrades FM coverage and accuracy performance.}
    \label{fig:ablation_cns_fm_autoencoder}
\end{figure}

The compression rate of the autoencoder reported in the main results is between 2x and 6x due to the original resolution being $64 \times 64$ (the rate varies due to the variation in the number of channels per dataset). While this compression rate is fairly small, it already had a detrimental effect on performance (coverage for \gls{fm} and accuracy for \gls{crps} trained models). In this ablation, we trained an autoencoder for a latent space at lower resolution of $8 \times 8 \times 8$ for \gls{cns}, increasing compression to 24x, while keeping the number of parameters roughly constant at $50$M. Figure~\ref{fig:ablation_cns_fm_autoencoder} shows the effect on \gls{fm} where coverage and accuracy degrade further at 24x compression, confirming that the latent space size is a limiting factor for performance in latent space models.

\clearpage
\subsection{Winkler-based monitoring}\label{appendix:ablations_winkler}

\begin{figure}[H]
    \centering
    \includegraphics[width=\linewidth]{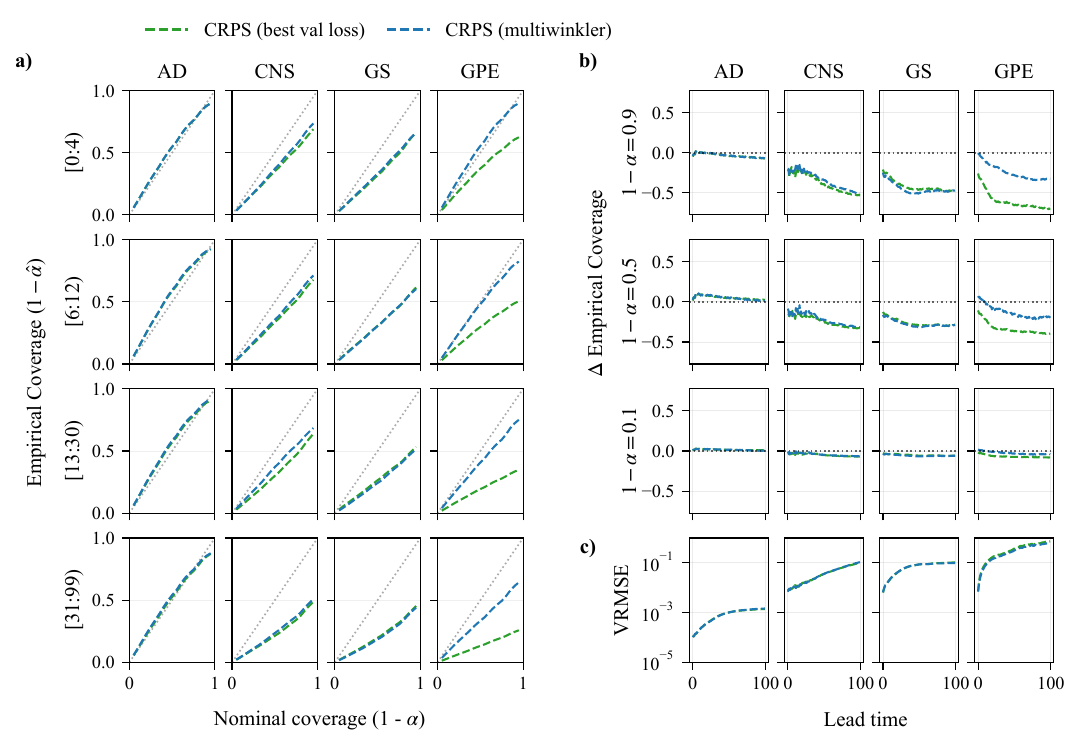}
    \caption{Checkpoint selection driven by Winkler score (default) compared with best validation loss ($M=8$) across datasets (\gls{ad}, \gls{cns}, \gls{gs}, \gls{gpe}). (a) Empirical vs.\ nominal coverage averaged over rollout windows. (b) $\Delta$ empirical coverage (observed minus nominal) by lead time over $100$-step rollouts. (c) \gls{vrmse} by lead time. The two checkpoint selection strategies have comparable coverage in most cases, but the Winkler score-driven strategy performs better on the \gls{gpe} dataset.}
    \label{fig:ablation_crps_multiwinkler}
\end{figure}

As discussed in Section \ref{sec:experimental_setup}, during \gls{crps} training we used the validation Winkler score to select the best model checkpoint. Figure \ref{fig:ablation_crps_multiwinkler} compares this against choosing a checkpoint based on the best validation loss. In most cases, the two checkpoint selection strategies have comparable coverage but the Winkler score checkpoint performs better on the \gls{gpe} dataset.

\clearpage
\subsection{CRPS loss variant}\label{appendix:ablations_crpss_loss}

\begin{figure}[H]
    \centering
    \includegraphics[width=\linewidth]{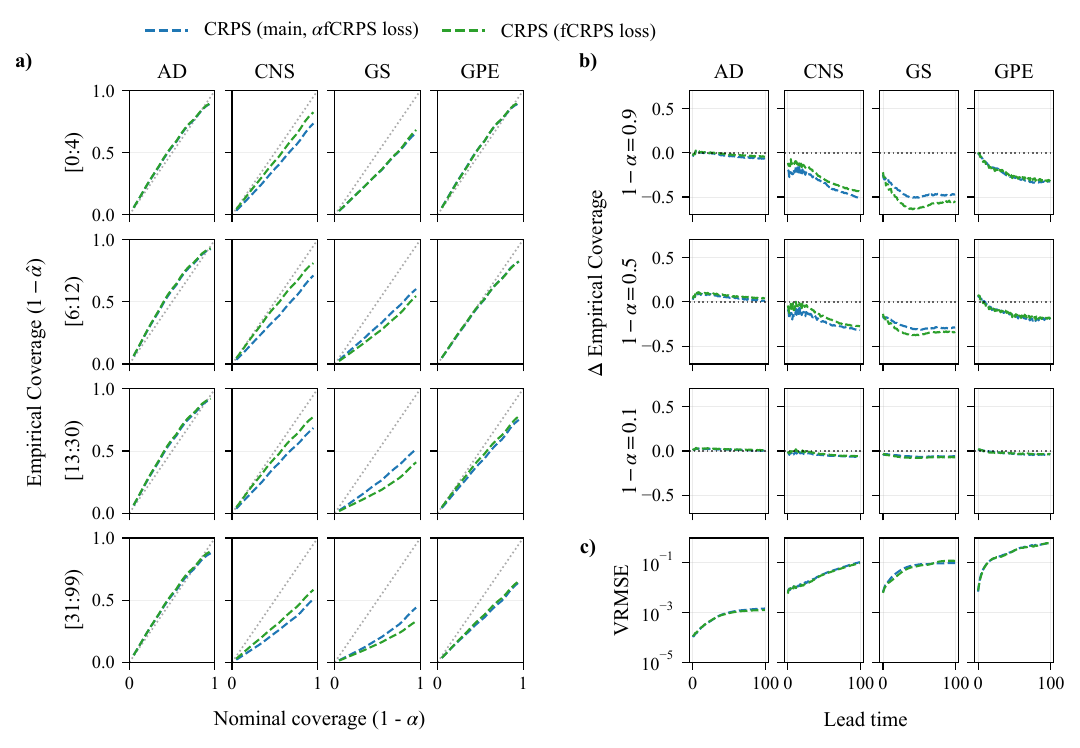}
    \caption{\gls{fcrps} vs.\ \gls{afcrps} training losses for the noise-injected ensemble ($M=8$) across datasets (\gls{ad}, \gls{cns}, \gls{gs}, \gls{gpe}). The main CRPS model uses \gls{afcrps} loss. (a) Empirical vs.\ nominal coverage averaged over rollout windows. (b) $\Delta$ empirical coverage (observed minus nominal) by lead time over $100$-step rollouts. (c) \gls{vrmse} by lead time.}
    \label{fig:ablation_fcrps_vs_main_crps}
\end{figure}

Across all four datasets, Figure~\ref{fig:ablation_fcrps_vs_main_crps} compares \gls{fcrps} with the \gls{afcrps} baseline from the main results. Rollout \gls{vrmse} remains similar. Coverage improves most clearly on \gls{cns}, changes little on \gls{ad} and \gls{gpe}, and worsens at long lead times on \gls{gs}.

\clearpage
\subsection{Ensemble size}\label{appendix:ablation_crps_ensemble_size}

\begin{figure}[H]
    \centering
    \includegraphics[width=\linewidth]{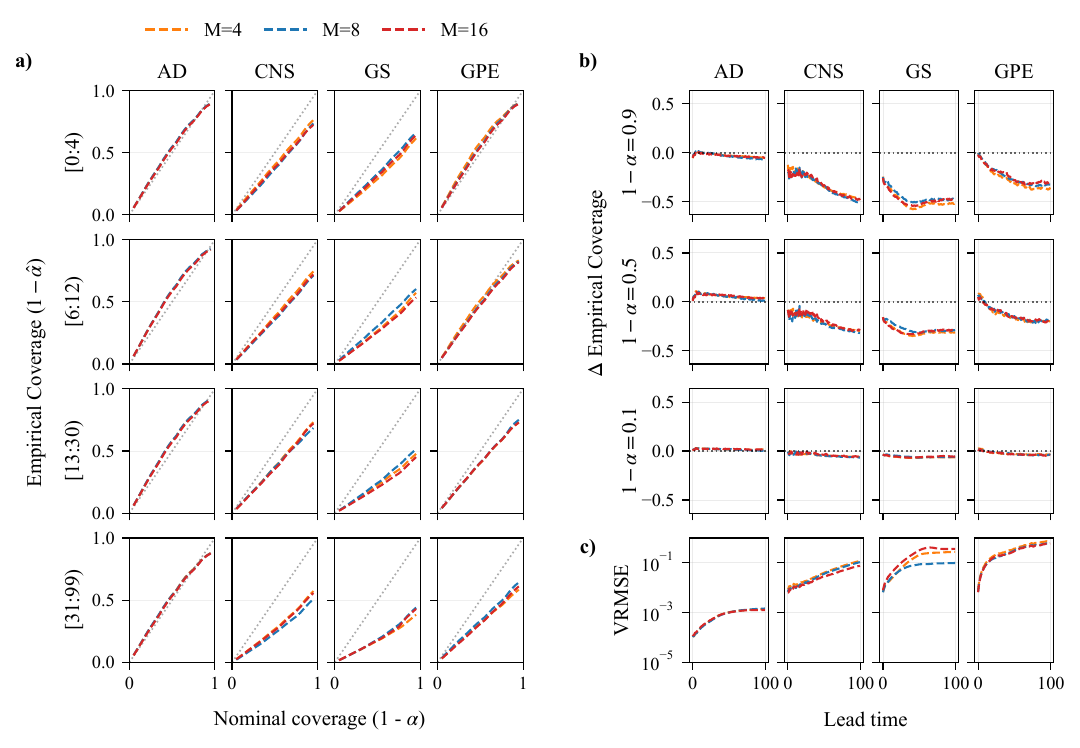}
    \caption{Effect of ensemble size on \gls{crps} training for $M\in\{4,8,16\}$ across datasets (\gls{ad}, \gls{cns}, \gls{gs}, \gls{gpe}). (a) Empirical vs.\ nominal coverage averaged over rollout windows. (b) $\Delta$ empirical coverage (observed minus nominal) vs.\ lead time over $100$-step rollouts. (c) \gls{vrmse} vs.\ lead time. Ensemble size does not have a consistent effect on coverage across datasets.}
    \label{fig:ablation_crps_ensemble_size}
\end{figure}

We compared the effect of ensemble size on coverage by comparing $M\in\{4,8,16\}$ on all datasets. As shown in Figure \ref{fig:ablation_crps_ensemble_size}, while there is some slight variability, we did not observe a consistent effect of ensemble size on coverage across datasets.

\clearpage
\subsection{Noise channels}\label{appendix:ablation_crps_noise_channels}

To create an ensemble for \gls{crps} training, we have to specify the number of noise channels used in the noise injection mechanism. As this increases the parameters in the model, we reduced the hidden dimension of the \gls{vit} backbone to maintain the same model size as used in \gls{fm}. In this ablation we evaluate the trade-offs between noise channel size and hidden dimension size by reducing the noise channels to 256 instead of the original 1024 and increasing the hidden dimension (from 568 to 704) to maintain the same model size.

\begin{figure}[H]
    \centering
    \includegraphics[width=0.9\linewidth]{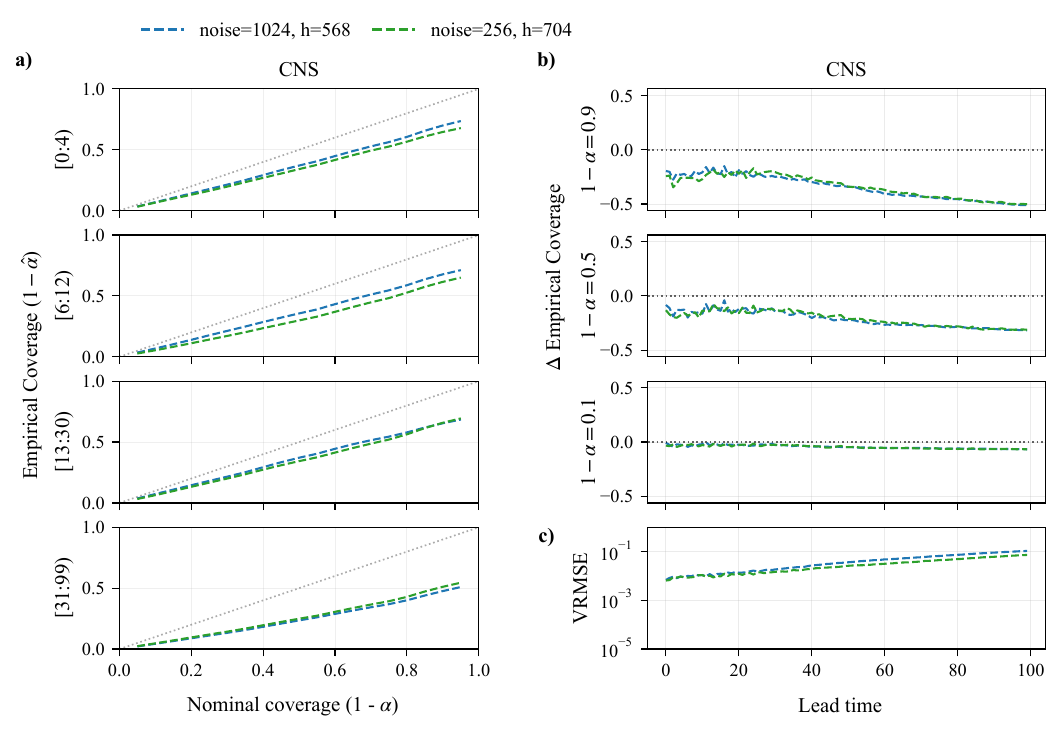}
    \caption{Effect of noise injection width (256 vs.\ 1024 noise channels) with adjusted hidden width to keep parameter count comparable on the \gls{cns} dataset. (a) Empirical vs.\ nominal coverage averaged over rollout windows. (b) $\Delta$ empirical coverage (observed minus nominal) by lead time over $100$-step rollouts. (c) \gls{vrmse} by lead time. The models perform comparably, although there may be a slight benefit from having larger noise channel size in terms of coverage.}
    \label{fig:ablation_cns_noise_channels}
\end{figure}

Figure \ref{fig:ablation_cns_noise_channels} shows that the model accuracy did not suffer from the reduced hidden dimension in the backbone, which is not surprising given the difference is fairly mild. Additionally, there might be a slight benefit from having larger noise channel size in terms of coverage.

\clearpage
\subsection{Alternative global conditioning strategies}\label{appendix:ablation_crps_global_conditioning}

For the \gls{crps}-trained \gls{vit} ensemble on \gls{cns} ($M=8$), we ablate how global simulator parameters are injected into the model. We compare (i) \textit{channel concatenation}, where a broadcast embedding of the conditioning variables is concatenated to the input channels, against (ii) \textit{backbone modulation}, where conditioning drives feature-wise modulation within the network (analogous to the conditioning mechanism used in our \gls{fm} backbone).

\begin{figure}[H]
    \centering
    \includegraphics[width=0.9\linewidth]{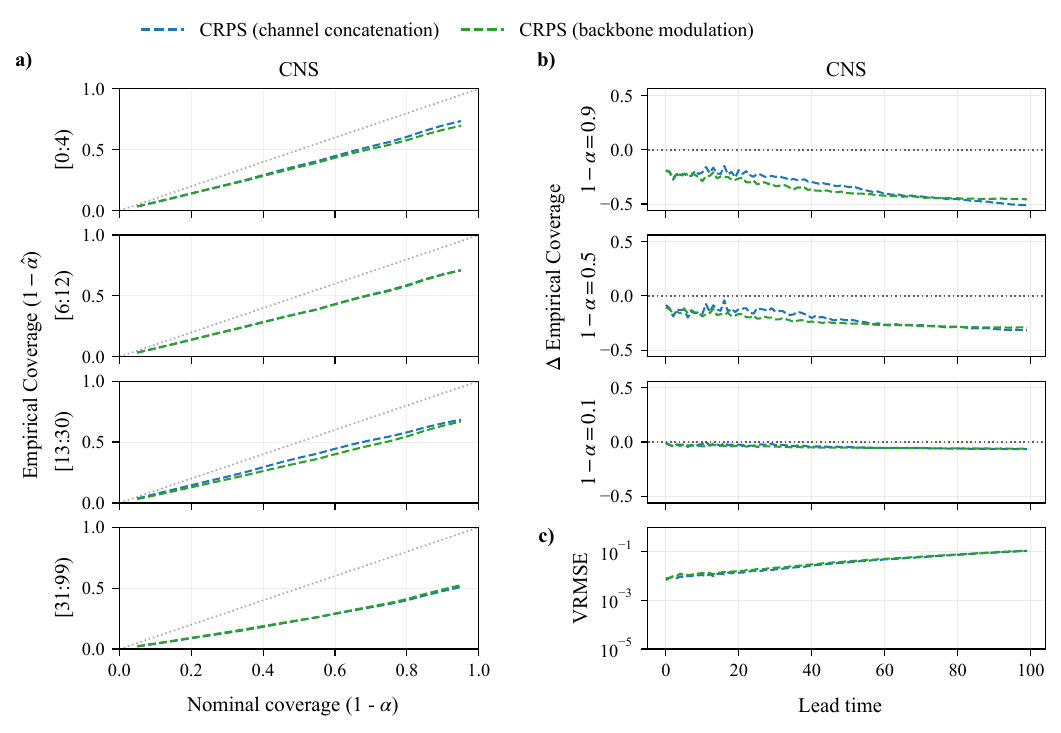}
    \caption{Global conditioning ablation for the \gls{crps}-trained \gls{vit} ensemble ($M=8$) on the \gls{cns} dataset. We compare conditioning via channel concatenation against backbone modulation. (a) Empirical vs.\ nominal coverage averaged over rollout windows. (b) $\Delta$ empirical coverage (observed minus nominal) by lead time over $100$-step rollouts. (c) \gls{vrmse} by lead time. Channel concatenation outperforms backbone modulation in terms of coverage performance.}
    \label{fig:ablation_cns_global_conditioning}
\end{figure}

Figure~\ref{fig:ablation_cns_global_conditioning} shows that our strategy of channel concatenation outperforms backbone modulation in terms of coverage performance.

\clearpage
\subsection{Flow matching ODE steps}\label{appendix:ablations_fm_ode_steps}

\begin{figure}[H]
    \centering
    \includegraphics[width=\linewidth]{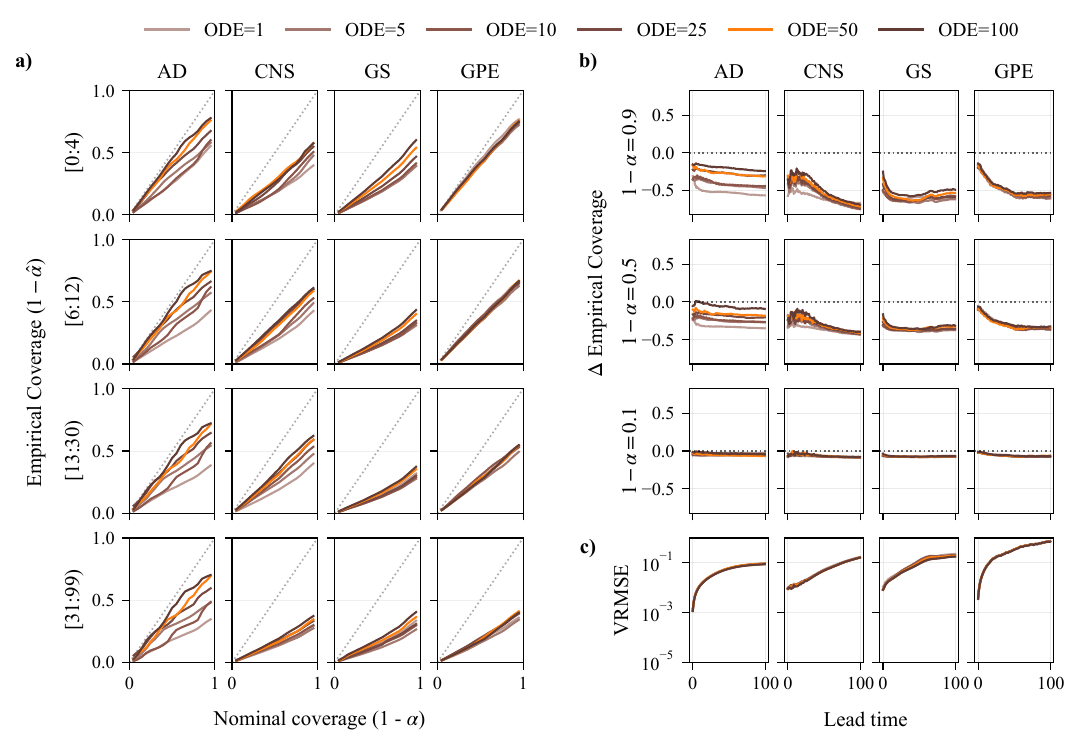}
    \caption{\gls{fm} performance as a function of the number of \gls{ode} integration steps used at sampling time across datasets (\gls{ad}, \gls{cns}, \gls{gs}, \gls{gpe}). (a) Empirical vs.\ nominal coverage averaged over rollout windows. (b) $\Delta$ empirical coverage (observed minus nominal) by lead time over $100$-step rollouts. (c) \gls{vrmse} by lead time. Coverage typically improves with increased \gls{ode} steps, while VRMSE remains similar.}
    \label{fig:ablation_fm_ode_steps}
\end{figure}

\gls{fm} sampling integrates a learned velocity field along a chosen number of \gls{ode} steps; more steps trade compute for sample quality. We swept the number of integration steps to explore performance sensitivity to this choice. Figure~\ref{fig:ablation_fm_ode_steps} shows that coverage typically improves with increased \gls{ode} steps, while VRMSE remains similar.

\clearpage
\subsection{Flow matching EMA weights}\label{appendix:ablations_fm_ema}

\begin{figure}[H]
    \centering
    \includegraphics[width=\linewidth]{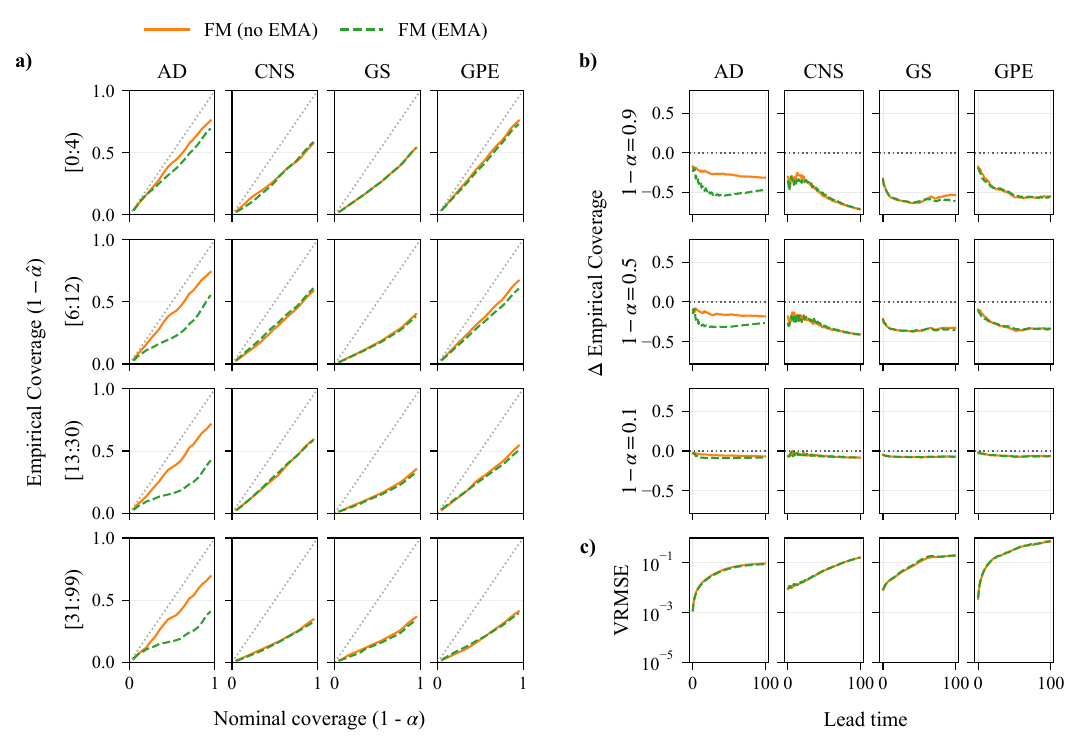}
    \caption{Original vs.\ EMA weights from the same selected latent \gls{fm} checkpoints across datasets (\gls{ad}, \gls{cns}, \gls{gs}, \gls{gpe}). Orange solid lines show the original weights used in the main results; green dashed lines show EMA weights. (a) Empirical vs.\ nominal coverage averaged over rollout windows. (b) $\Delta$ empirical coverage (observed minus nominal) by lead time over $100$-step rollouts. (c) \gls{vrmse} by lead time.}
    \label{fig:ablation_fm_ema}
\end{figure}

For the latent \gls{fm} models, we compare the original weights used in the main results with their exponential moving average (EMA, decay $0.999$) from the same selected checkpoint. Both evaluations use the same autoencoder, $10$ ensemble members and $50$ Euler steps. Figure~\ref{fig:ablation_fm_ema} shows no consistent benefit from EMA: coverage worsens most clearly on \gls{ad}, with smaller differences on the other datasets. Rollout \gls{vrmse} remains broadly similar, with some increase on \gls{gs}.

\clearpage
\subsection{Alternative processor architectures}\label{appendix:ablation_architectures}

We tested whether the results generalise across different architectures. Specifically, we compared the \gls{fm} model against denoising diffusion \citep{ho2020} using the same \gls{vit} backbone. We also compared a \gls{crps}-trained ensemble of U-Net models against the \gls{vit} ensemble. We used a U-Net implemented in \texttt{Azula} \citep{azula}. This is an adaptation of the classic architecture from \citet{unet} to work as a backbone in generative models. Using this variant allowed us to use a similar noise injection strategy as that used in the \gls{vit} ensemble (see Appendix~\ref{appendix:model_archs}). All models were matched on size and evaluated using the metrics described in Section \ref{sec:experimental_setup}.

\begin{figure}[H]
    \centering
    \includegraphics[width=\linewidth]{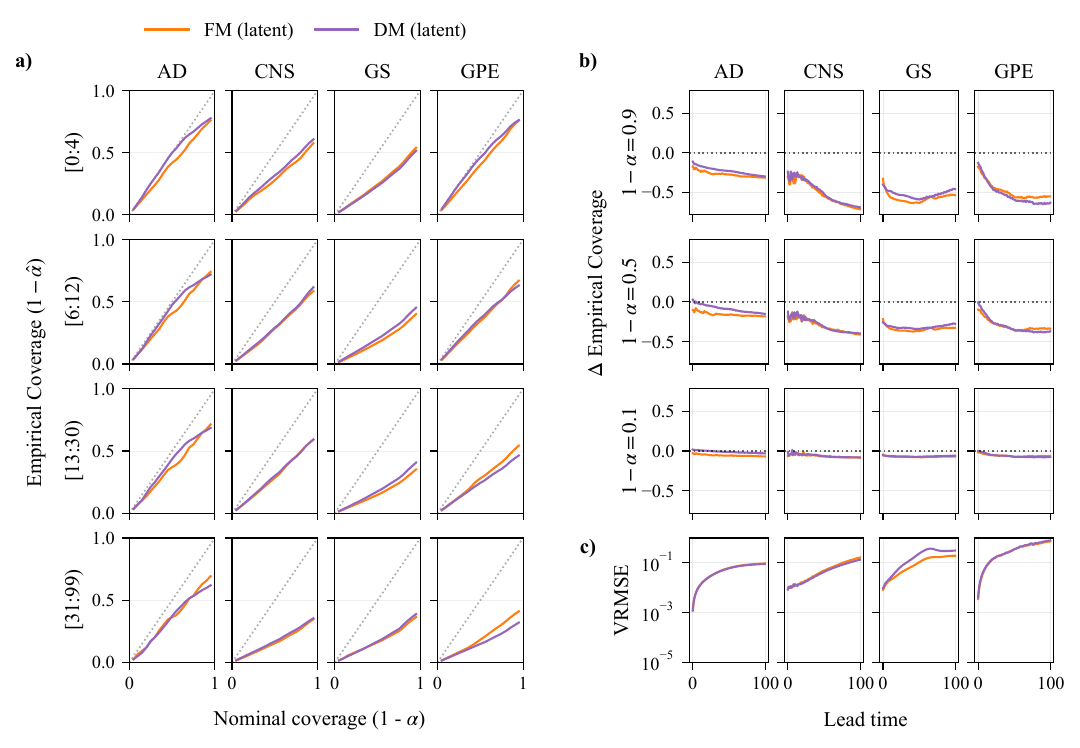}
    \caption{\gls{fm} vs.\ denoising diffusion in \emph{latent} space (matched backbones) across datasets (\gls{ad}, \gls{cns}, \gls{gs}, \gls{gpe}). (a) Empirical vs.\ nominal coverage averaged over rollout windows. (b) $\Delta$ empirical coverage (observed minus nominal) by lead time over $100$-step rollouts. (c) \gls{vrmse} by lead time.}
    \label{fig:ablation_dm_latent}
\end{figure}

\begin{figure}[H]
    \centering
    \includegraphics[width=\linewidth]{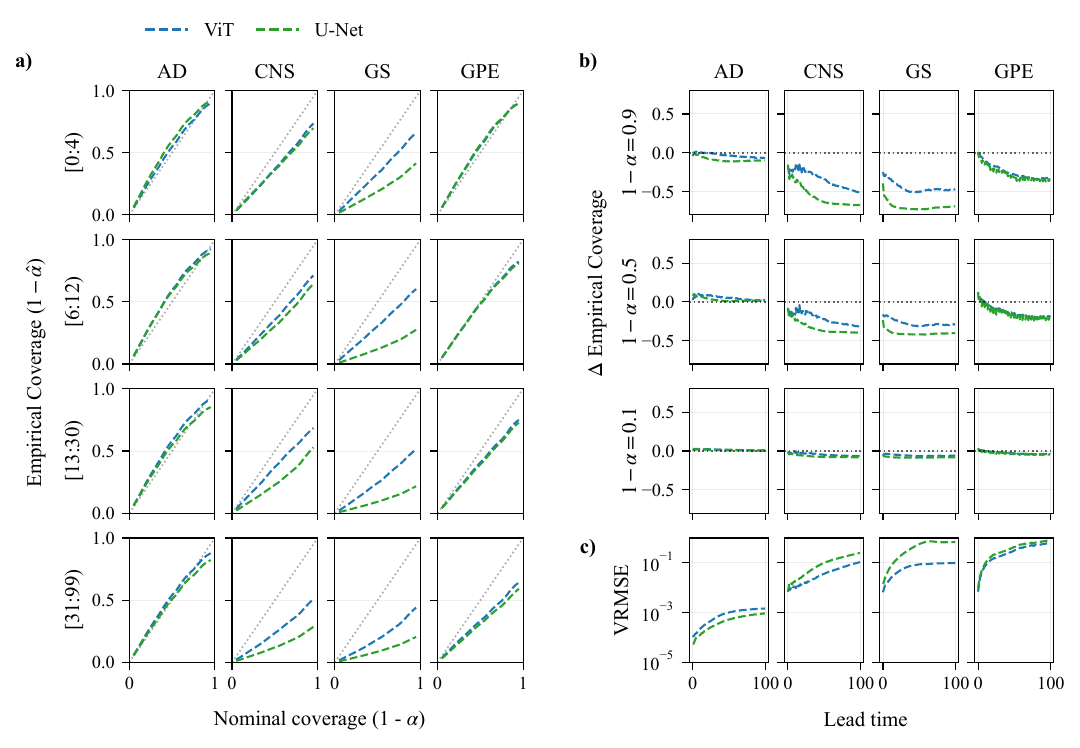}
    \caption{\gls{vit} vs.\ U-Net \gls{crps}-trained ensemble ($M=8$) performance across datasets (\gls{ad}, \gls{cns}, \gls{gs}, \gls{gpe}). (a) Empirical vs.\ nominal coverage averaged over rollout windows. (b) $\Delta$ empirical coverage (observed minus nominal) vs.\ lead time over $100$-step rollouts. (c) \gls{vrmse} vs.\ lead time.}
    \label{fig:ablation_vit_unet}
\end{figure}

Figures \ref{fig:ablation_dm_latent} and \ref{fig:ablation_vit_unet} compare these models across all four datasets (\gls{ad}, \gls{cns}, \gls{gs}, \gls{gpe}). Diffusion and \gls{fm} have broadly similar rollout \gls{vrmse} on \gls{ad}, \gls{cns} and \gls{gpe}, while diffusion has higher rollout error on \gls{gs}; coverage differences vary with dataset and lead time. The U-Net ensemble has lower \gls{vrmse} on \gls{ad}, but higher error on \gls{cns}, \gls{gs} and \gls{gpe} and worse rollout coverage on \gls{cns} and \gls{gs} than the \gls{vit} ensemble.

\clearpage
\subsection{Fourier neural operator backbone}\label{appendix:fno_backbone}

We evaluate how the \gls{crps}-training framework performs with a Fourier neural operator (FNO) backbone~\citep{fno}. An FNO layer combines a pointwise linear map with a learned spectral convolution,
\begin{equation}
    \bm{h}_{j+1}(\bm{s})
    =\sigma\!\left(
        W_j\bm{h}_j(\bm{s})
        +\mathcal{F}^{-1}\!\left[
            R_j\odot\mathcal{F}[\bm{h}_j]
        \right](\bm{s})
    \right),
\end{equation}
where $\bm{s}$ is spatial position, $\bm{h}_j(\bm{s})\in\mathbb{R}^{d}$ is the activation entering layer $j$, $\sigma$ is an elementwise nonlinear activation, and $j\in\{0,\ldots,3\}$ indexes the four FNO layers; here $d=264$. The matrix $W_j$ mixes channels pointwise, $\mathcal{F}$ is the spatial Fourier transform, and $R_j$ contains learned multipliers for the retained modes. Both $\mathcal{F}[\bm{h}_j]$ and $R_j$ are complex valued, so the full complex coefficients, including their magnitude and phase, contribute to the prediction. The direct spectral path truncates modes outside the retained band, although the pointwise path and nonlinear layers can still generate higher-frequency structure~\citep{fno}.

Our FNO uses four layers, $264$ hidden channels, and $16$ modes along each spatial axis. Counting each complex coefficient as two real scalars gives approximately $81.1$M parameters, closely matching the $80.8$M ViT baseline. Because the FNO does not use the ViT conditional-normalisation path, each of the eight ensemble members instead receives an independently sampled spatial white-noise channel concatenated to its input. Both backbones use ambient-space \gls{afcrps} training, the same optimiser and training budget, and the same checkpoint-selection and evaluation protocol. FourCastNet~3 uses a related spherical neural operator for probabilistic weather forecasting, but combines local spherical convolutions with global spectral convolutions defined through spherical harmonic transforms to respect the geometry of the globe, rather than using the planar Fourier transform employed here~\citep{sfno,fourcastnet}.

\begin{figure}[H]
    \centering
    \includegraphics[width=\linewidth]{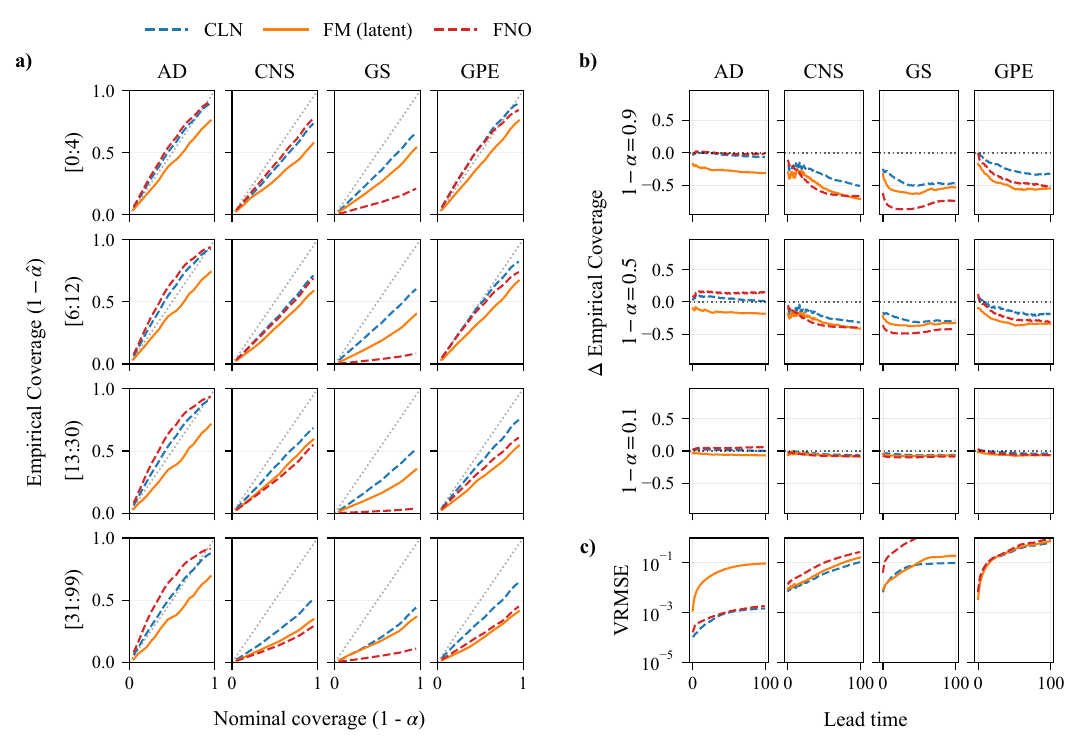}
    \caption{FNO and ViT \gls{crps}-trained ensembles (the ViT ensemble is labelled CLN), with latent \gls{fm} included for context, across datasets (\gls{ad}, \gls{cns}, \gls{gs}, \gls{gpe}). (a) Empirical vs.\ nominal coverage averaged over rollout windows. (b) $\Delta$ empirical coverage (observed minus nominal) by lead time over $100$-step rollouts. (c) \gls{vrmse} by lead time. The FNO is competitive on some early rollout windows but degrades more consistently over long rollouts.}
    \label{fig:ablation_fno}
\end{figure}

\begin{table}[H]
    \centering
    \caption{Single-step comparison of the ViT and FNO backbones under \gls{crps} training. Bold denotes the better result for each dataset.}
    \label{tab:ablation_fno}
    \small
\setlength{\tabcolsep}{4pt}
\begin{tabular}{@{}llcccc@{}}
\toprule
Dataset & Backbone & VRMSE $\downarrow$ & Coverage MAE $\downarrow$ & CRPS $\downarrow$ & SSR $\to 1$ \\
\midrule
AD  & ViT & 1.8e-04 & \bfseries 0.04 & \bfseries 8.8e-05 & \bfseries 1.07 \\
    & FNO & \bfseries 1.7e-04 & 0.06 & 8.9e-05 & 1.82 \\
\midrule
CNS & ViT & \bfseries 5.6e-03 & \bfseries 0.12 & \bfseries 2.7e-03 & \bfseries 1.16 \\
    & FNO & 1.4e-02 & 0.18 & 6.4e-03 & 0.71 \\
\midrule
GS  & ViT & \bfseries 1.1e-03 & 0.20 & \bfseries 5.2e-04 & \bfseries 0.67 \\
    & FNO & 3.7e-03 & \bfseries 0.18 & 2.3e-03 & 0.56 \\
\midrule
GPE & ViT & \bfseries 2.8e-02 & \bfseries 0.04 & \bfseries 9.7e-03 & \bfseries 0.92 \\
    & FNO & 6.3e-02 & 0.09 & 2.4e-02 & 0.56 \\
\bottomrule
\end{tabular}

\end{table}

The FNO is competitive for single-step \gls{ad} accuracy and \gls{gs} coverage, but the ViT has lower coverage MAE on the other systems and better \gls{ssr} on all four (Table~\ref{tab:ablation_fno}). Figure~\ref{fig:ablation_fno} shows a clearer rollout separation with the ViT backbone providing more consistent rollout accuracy and coverage, with lower coverage MAE in the $[31{:}99)$ window on all four datasets. The largest long-window \gls{vrmse} gap occurs on \gls{gs} ($1.3$ for FNO versus $0.095$ for ViT), apparently reflecting spatial displacement of otherwise coherent localised structures.

\clearpage
\raggedbottom
\subsection{Sensitivity to training seed}\label{appendix:training_seed}

We retrained every model in the main comparison with an independent random seed while keeping the training, validation, and test trajectories and all other training settings fixed. The comparison measures sensitivity to optimisation and other seeded training operations.

\begin{table}[H]
    \centering
    \caption{Single-step results for the original and repeated training runs, reported as seed 1 / seed 2. Both seeds use the same training, validation, and test trajectories.}
    \label{tab:ablation_training_seed}
    \small
\setlength{\tabcolsep}{4pt}
\begin{tabular}{@{}llcccc@{}}
\toprule
Dataset & Model & VRMSE $\downarrow$ & Coverage MAE $\downarrow$ & CRPS $\downarrow$ & SSR $\to 1$ \\
\midrule
AD  & CRPS & 1.8e-04 / 1.9e-04 & 0.04 / 0.04 & 8.8e-05 / 9.2e-05 & 1.07 / 1.06 \\
    & FM   & 2.2e-04 / 2.1e-04 & 0.02 / 0.02 & 1.3e-04 / 1.2e-04 & 1.31 / 1.27 \\
\midrule
CNS & CRPS & 5.6e-03 / 5.8e-03 & 0.12 / 0.10 & 2.7e-03 / 2.8e-03 & 1.16 / 1.18 \\
    & FM   & 8.7e-03 / 8.7e-03 & 0.28 / 0.29 & 3.9e-03 / 3.9e-03 & 0.65 / 0.63 \\
\midrule
GS  & CRPS & 1.1e-03 / 1.1e-03 & 0.20 / 0.21 & 5.2e-04 / 5.2e-04 & 0.67 / 0.65 \\
    & FM   & 1.3e-03 / 1.3e-03 & 0.18 / 0.18 & 5.1e-04 / 5.0e-04 & 0.57 / 0.59 \\
\midrule
GPE & CRPS & 2.8e-02 / 3.0e-02 & 0.04 / 0.04 & 9.7e-03 / 1.1e-02 & 0.92 / 0.96 \\
    & FM   & 3.2e-02 / 2.8e-02 & 0.24 / 0.23 & 1.1e-02 / 9.6e-03 & 0.44 / 0.48 \\
\bottomrule
\end{tabular}

\end{table}

\begin{figure}[H]
    \centering
    \includegraphics[width=0.82\linewidth]{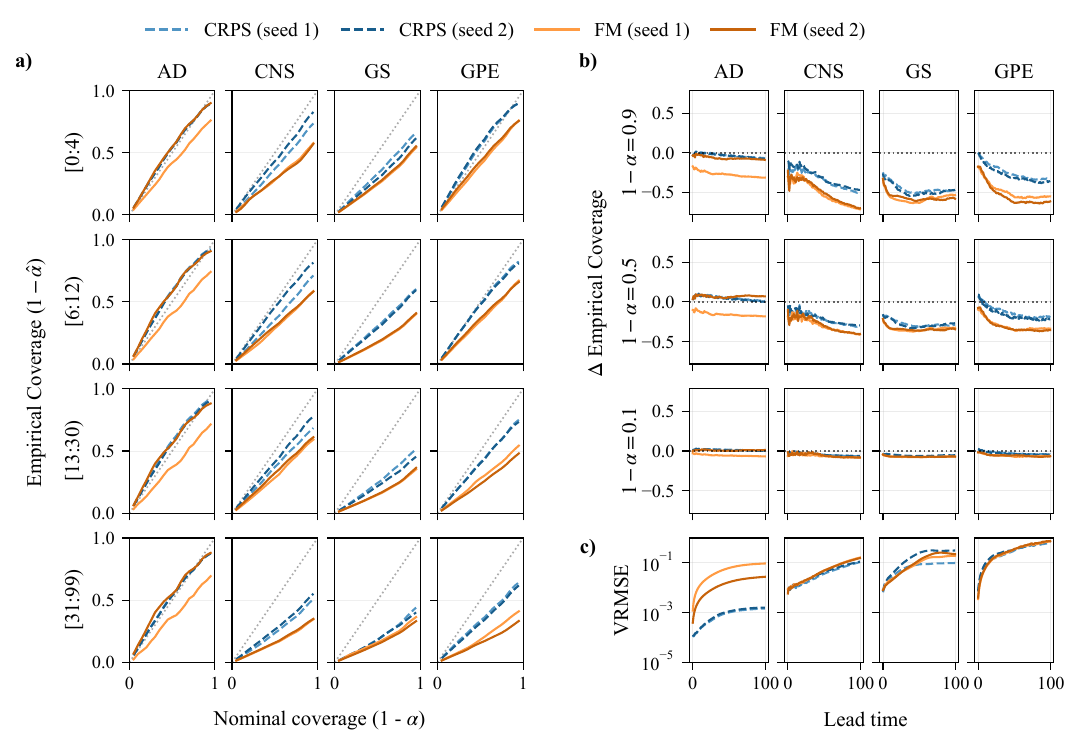}
    \caption{Performance of the main \gls{crps} and \gls{fm} models across datasets (\gls{ad}, \gls{cns}, \gls{gs}, \gls{gpe}) reported for two independent training runs with different random seeds, denoted as seed 1 / seed 2. (a) Empirical vs.\ nominal coverage averaged over rollout windows. (b) $\Delta$ empirical coverage (observed minus nominal) by lead time over $100$-step rollouts. (c) \gls{vrmse} by lead time. The two seeds give similar overall coverage trends but \gls{fm} coverage and rollout \gls{vrmse} differ notably for \gls{ad}. Seed differences are also visible in \gls{crps} coverage on \gls{cns} and rollout \gls{vrmse} on \gls{gs}.}
    \label{fig:ablation_training_seed}
\end{figure}

Table~\ref{tab:ablation_training_seed} shows close agreement in the single-step metrics across both seeds. Coverage MAE and \gls{ssr} retain the same qualitative pattern on all four systems. For seed 2 on \gls{gpe}, \gls{fm} has lower \gls{vrmse} and \gls{crps}, but the \gls{crps}-trained ensemble still has substantially lower coverage MAE and an \gls{ssr} closer to one. Figure~\ref{fig:ablation_training_seed} shows that the full rollout trajectories are also broadly consistent, although seed sensitivity is more visible for \gls{fm} accuracy on \gls{ad} and \gls{crps} accuracy on \gls{gs} at long lead times.

\subsection{Sensitivity to training data}\label{appendix:training_data}

The seed experiment conditions on a fixed benchmark. We separately test sensitivity to the finite training sample on all four datasets by generating independent training trajectories from the same simulator parameter distributions and retraining both pipelines. The original test trajectories, training seed, and all other training settings remain fixed, so each main model and training-data rerun is evaluated on the same test cases. This isolates the effect of replacing the sampled training trajectories, following the fixed-test design used in related PDE-emulator datasets~\citep{pdebench,thewell}.

\begin{table}[H]
    \centering
    \caption{Single-step results for the main training data and the independently generated training set, reported as main / new training data. All pairs use the same test trajectories; the corresponding rollout estimates appear in Figure~\ref{fig:ablation_training_data}.}
    \label{tab:ablation_training_data}
    \footnotesize
\setlength{\tabcolsep}{2.5pt}
\begin{tabular}{@{}llcccc@{}}
\toprule
Dataset & Model & VRMSE $\downarrow$ & Coverage MAE $\downarrow$ & CRPS $\downarrow$ & SSR $\to 1$ \\
\midrule
AD  & CRPS & 1.8e-04 / 1.7e-04 & 0.041 / 0.040 & 8.8e-05 / 8.3e-05 & 1.07 / 1.08 \\
    & FM   & 2.2e-04 / 2.1e-04 & 0.019 / 0.019 & 1.3e-04 / 1.2e-04 & 1.30 / 1.30 \\
\midrule
CNS & CRPS & 5.6e-03 / 6.0e-03 & 0.124 / 0.111 & 2.7e-03 / 2.9e-03 & 1.16 / 1.11 \\
    & FM   & 8.7e-03 / 8.9e-03 & 0.284 / 0.288 & 3.9e-03 / 4.1e-03 & 0.66 / 0.64 \\
\midrule
GS  & CRPS & 1.1e-03 / 1.2e-03 & 0.198 / 0.20 & 5.2e-04 / 6.1e-04 & 0.67 / 0.66 \\
    & FM   & 1.3e-03 / 1.3e-03 & 0.183 / 0.186 & 5.1e-04 / 5.2e-04 & 0.57 / 0.58 \\
\midrule
GPE & CRPS & 2.8e-02 / 3.0e-02 & 0.034 / 0.026 & 9.6e-03 / 1.1e-02 & 0.92 / 0.99 \\
    & FM   & 3.2e-02 / 3.0e-02 & 0.24 / 0.24 & 1.1e-02 / 1.0e-02 & 0.43 / 0.45 \\
\bottomrule
\end{tabular}

\end{table}

\begin{figure}[H]
    \centering
    \includegraphics[width=0.82\linewidth]{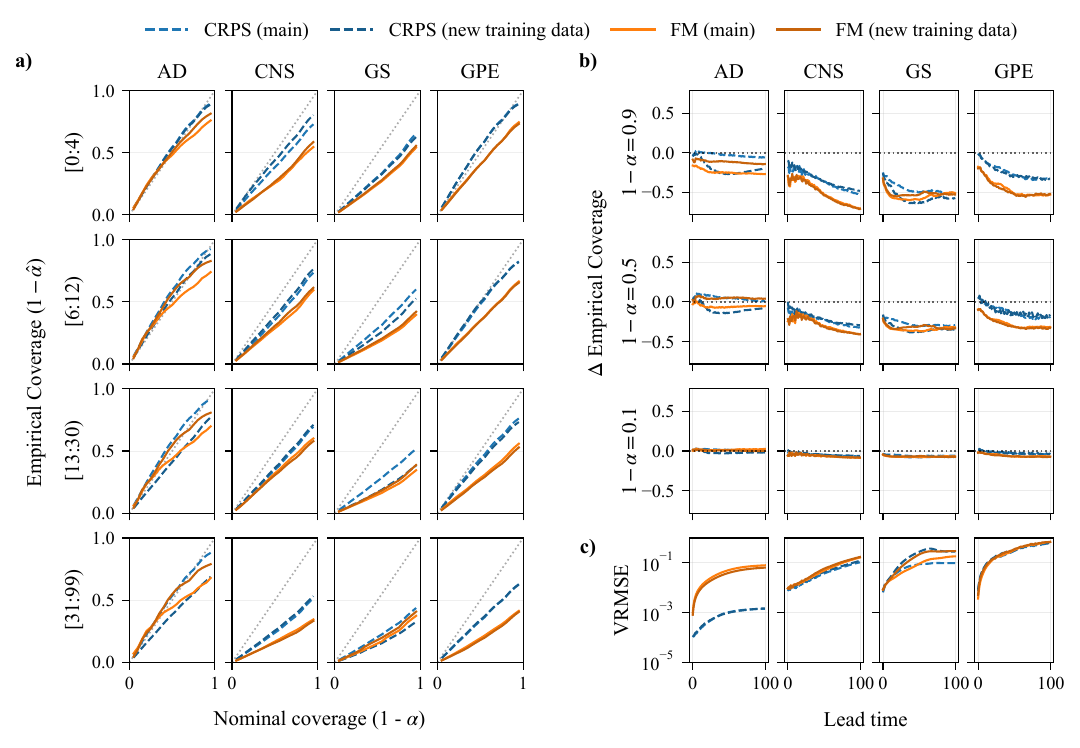}
    \caption{Training-data sensitivity of the main \gls{crps} and \gls{fm} models across datasets (\gls{ad}, \gls{cns}, \gls{gs}, \gls{gpe}). Each rerun uses a new, independently generated training set but the same test trajectories as the main models. (a) Empirical vs.\ nominal coverage averaged over rollout windows. (b) $\Delta$ empirical coverage (observed minus nominal) by lead time over $100$-step rollouts. (c) \gls{vrmse} by lead time.}
    \label{fig:ablation_training_data}
\end{figure}

The paired single-step results in Table~\ref{tab:ablation_training_data} are close across all four datasets, and the independently trained models track their main-run counterparts over much of the rollout in Figure~\ref{fig:ablation_training_data}. For \gls{crps} on \gls{cns}, coverage MAE decreases from $0.124$ to $0.111$, while \gls{vrmse} and \gls{crps} increase slightly; the \gls{fm} single-step metrics also change only modestly across the four systems. Differences become more visible in some long-horizon curves, particularly \gls{gs} \gls{vrmse}, so individual rollout trajectories are not invariant to the sampled training set. Nevertheless, the broad calibration and accuracy comparison between \gls{crps} and \gls{fm} remains consistent across the four controlled reruns.

\subsection{MC dropout trained with CRPS loss}\label{appendix:mc_dropout_crps}

The main \gls{crps}-trained ensemble samples Gaussian noise and injects it through \gls{cln}. In this ablation, we replace that stochastic modulation with dropout in the feed-forward blocks of the ViT backbone. Specifically, an elementwise Bernoulli mask is applied after the activation and before the second linear map in the feed-forward branch of every transformer block. Each feed-forward network expands the token width from $d=704$ to $4d=2816$ before the activation. For a processor batch of size $B$ and $L=16\times16=256$ spatial patch tokens, the corresponding mask in block $j$ is therefore $\bm{\xi}_j\in\{0,1\}^{B\times L\times 4d}$; $\bm{\xi}=(\bm{\xi}_1,\ldots,\bm{\xi}_{12})$ denotes the collection for one processor evaluation. Dropout is applied with probability $p=0.1$, remains active at evaluation time, and $\bm{\xi}$ is resampled independently on every forward call and at every autoregressive rollout step. Removing the $1024$-dimensional modulation path reduces the processor size, so we increase the ViT width from $568$ to $704$ while retaining $12$ blocks and $8$ heads, giving $79.8$M parameters compared with $80.8$M for the main ViT.

Training otherwise follows the ambient-space \gls{crps} protocol used throughout: eight predictions with independently sampled mask collections are evaluated for each of $32$ cases per GPU update and optimised with \gls{afcrps}. Evaluation uses ten stochastic forward passes. This treats \gls{mc} dropout as a shared-parameter ensemble sampler~\citep{gal2016}; closely related weather-forecasting work has likewise trained dropout samples using a \gls{crps}-family objective~\citep{ucast, otterweather}.

\begin{figure}[H]
    \centering
    \includegraphics[width=\linewidth]{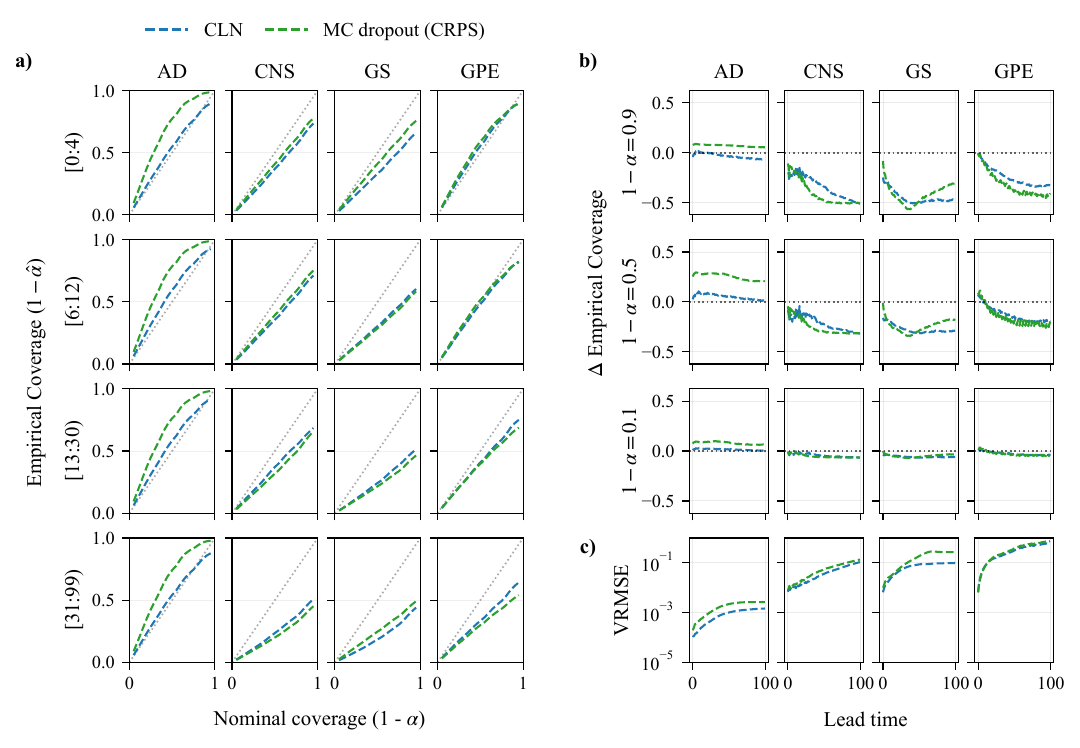}
    \caption{\gls{cln} vs.\ \gls{mc} dropout as the stochastic mechanism for \gls{crps}-trained ensembles across datasets (\gls{ad}, \gls{cns}, \gls{gs}, \gls{gpe}). (a) Empirical vs.\ nominal coverage averaged over rollout windows. (b) $\Delta$ empirical coverage (observed minus nominal) by lead time over $100$-step rollouts. (c) \gls{vrmse} by lead time. \gls{mc} dropout changes coverage in a dataset-dependent manner.}
    \label{fig:ablation_mc_dropout_crps}
\end{figure}

\begin{table}[H]
    \centering
    \caption{Single-step comparison of \gls{cln} and \gls{mc} dropout as ensemble mechanisms under \gls{crps} training. Bold denotes the better result for each dataset.}
    \label{tab:ablation_mc_dropout_crps}
    \small
\setlength{\tabcolsep}{4pt}
\begin{tabular}{@{}llcccc@{}}
\toprule
Dataset & Mechanism & VRMSE $\downarrow$ & Coverage MAE $\downarrow$ & CRPS $\downarrow$ & SSR $\to 1$ \\
\midrule
AD  & CLN               & \bfseries 1.8e-04 & \bfseries 0.04 & \bfseries 8.8e-05 & \bfseries 1.07 \\
    & MC dropout (CRPS) & 2.8e-04 & 0.15 & 1.8e-04 & 2.16 \\
\midrule
CNS & CLN               & \bfseries 5.6e-03 & 0.12 & \bfseries 2.7e-03 & 1.16 \\
    & MC dropout (CRPS) & 6.2e-03 & \bfseries 0.09 & 3.0e-03 & \bfseries 1.10 \\
\midrule
GS  & CLN               & \bfseries 1.1e-03 & 0.20 & \bfseries 5.2e-04 & \bfseries 0.67 \\
    & MC dropout (CRPS) & 1.4e-03 & \bfseries 0.04 & 6.3e-04 & 1.55 \\
\midrule
GPE & CLN               & \bfseries 2.8e-02 & 0.04 & \bfseries 9.7e-03 & \bfseries 0.92 \\
    & MC dropout (CRPS) & 3.2e-02 & \bfseries 0.03 & 1.2e-02 & 0.84 \\
\bottomrule
\end{tabular}

\end{table}

Table~\ref{tab:ablation_mc_dropout_crps} and Figure~\ref{fig:ablation_mc_dropout_crps} show that neither stochastic mechanism dominates coverage across datasets and lead times, whereas \gls{cln} has lower single-step \gls{vrmse} and \gls{crps} on all four datasets and generally retains lower rollout \gls{vrmse}. For single-step prediction, \gls{mc} dropout improves coverage MAE on \gls{gs} and worsens it on \gls{ad}, with smaller improvements on \gls{cns} and \gls{gpe}. A broader hyperparameter study would be needed to determine whether either mechanism is generally preferable.

\subsection{MC dropout trained with MSE loss}\label{appendix:mc_dropout_mse}

We evaluate conventional \gls{mc} dropout trained with $L_2$-regularised MSE loss using the same parameter-matched ViT as in Appendix~\ref{appendix:mc_dropout_crps}. Dropout remains restricted to the feed-forward branches with $p=0.1$, using the mask collection $\bm{\xi}$ defined above, but training uses one independently sampled $\bm{\xi}$ per case. The per-GPU batch size increases from $32$ to $256$, keeping $256$ processor evaluations per update. Using the processor $f_\rho$ and its parameter collection $\rho$ defined in Section~\ref{sec:methodology:ensemble}, let $\rho_{\geq 2}\subset\rho$ contain the trainable processor parameters with at least two dimensions. The training objective is
\begin{equation}
    \mathcal{L}_{\mathrm{MSE+L2}}(\rho;\bm{\xi})
    =\operatorname{MSE}\!\left(
        f_{\rho}(\bm{x}_{\mathrm{in}},\bm{c};\bm{\xi}),
        \bm{x}_{\mathrm{out}}
    \right)
    +\lambda\sum_{P\in\rho_{\geq 2}}\lVert P\rVert_{F}^{2},
    \qquad \lambda=10^{-5}.
\end{equation}
For this conventional \gls{mc}-dropout baseline, we adopt dropout probability $p=0.1$ and regularisation coefficient $\lambda=10^{-5}$, matching the released WeatherBench Probability configuration~\citep{garg2022}. Biases and normalisation vectors are excluded from the penalty. AdamW weight decay is zero, and the $L_2$ penalty is applied only during training; validation and test loss use plain MSE. At evaluation, ten stochastic forward passes form the predictive ensemble.

\begin{table}[H]
    \centering
    \caption{Single-step comparison of the main \gls{crps} and latent \gls{fm} models with conventional \gls{mc} dropout trained using MSE and explicit $L_2$ regularisation. Bold denotes the best result for each dataset.}
    \label{tab:ablation_mc_dropout_mse}
    \small
\setlength{\tabcolsep}{4pt}
\begin{tabular}{@{}llcccc@{}}
\toprule
Dataset & Model & VRMSE $\downarrow$ & Coverage MAE $\downarrow$ & CRPS $\downarrow$ & SSR $\to 1$ \\
\midrule
AD  & CRPS             & \bfseries 1.8e-04 & 0.04 & \bfseries 8.8e-05 & 1.07 \\
    & FM (latent)      & 2.2e-04 & \bfseries 0.02 & 1.3e-04 & 1.31 \\
    & MC dropout (MSE) & 4.8e-03 & 0.07 & 2.6e-03 & \bfseries 1.05 \\
\midrule
CNS & CRPS             & \bfseries 5.6e-03 & \bfseries 0.12 & \bfseries 2.7e-03 & \bfseries 1.16 \\
    & FM (latent)      & 8.7e-03 & 0.28 & 3.9e-03 & 0.65 \\
    & MC dropout (MSE) & 1.9e-02 & 0.19 & 9.2e-03 & 0.68 \\
\midrule
GS  & CRPS             & \bfseries 1.1e-03 & 0.20 & 5.2e-04 & 0.67 \\
    & FM (latent)      & 1.3e-03 & 0.18 & \bfseries 5.1e-04 & 0.57 \\
    & MC dropout (MSE) & 1.1e-02 & \bfseries 0.06 & 3.9e-03 & \bfseries 0.93 \\
\midrule
GPE & CRPS             & \bfseries 2.8e-02 & \bfseries 0.04 & \bfseries 9.7e-03 & 0.92 \\
    & FM (latent)      & 3.2e-02 & 0.24 & 1.1e-02 & 0.44 \\
    & MC dropout (MSE) & 3.6e-02 & \bfseries 0.04 & 1.6e-02 & \bfseries 0.98 \\
\bottomrule
\end{tabular}

\end{table}

Conventional \gls{mc} dropout has the highest single-step \gls{vrmse} and \gls{crps} on every dataset (Table~\ref{tab:ablation_mc_dropout_mse}). Its competitive coverage MAE on \gls{gs} and \gls{gpe} does not imply accurate or sharp predictions.

\begin{figure}[H]
    \centering
    \includegraphics[width=0.80\linewidth]{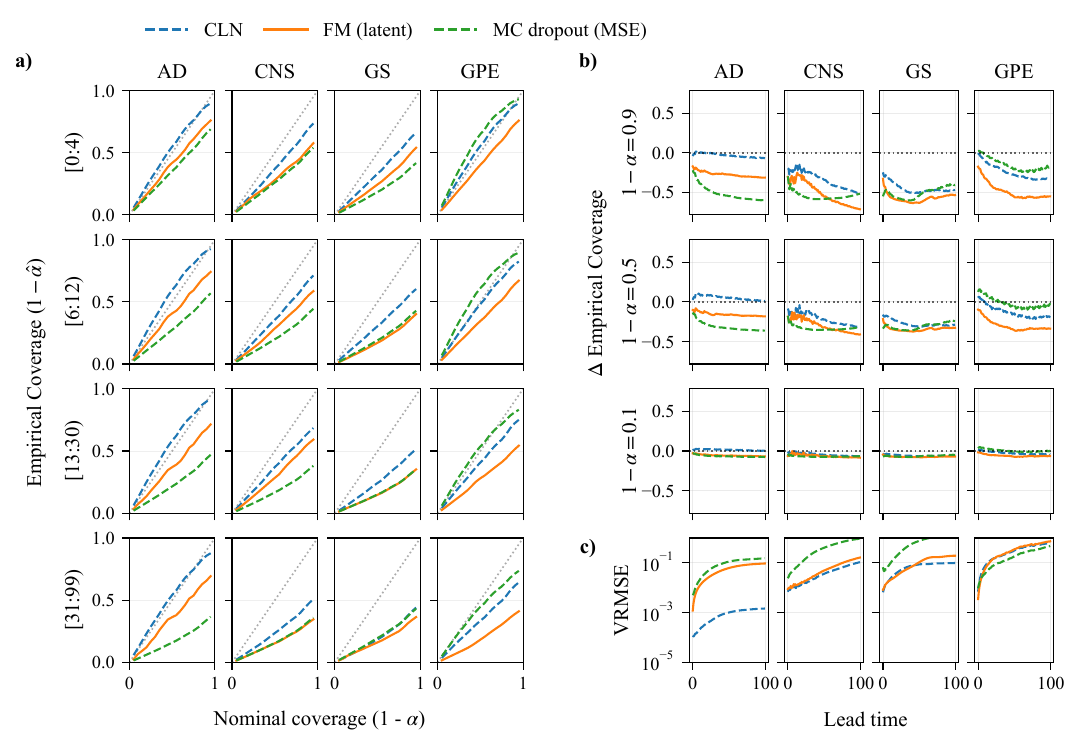}
    \caption{Conventional \gls{mc} dropout trained with MSE and explicit $L_2$ regularisation compared with the main \gls{crps} and latent \gls{fm} models across datasets (\gls{ad}, \gls{cns}, \gls{gs}, \gls{gpe}). (a) Empirical vs.\ nominal coverage averaged over rollout windows. (b) $\Delta$ empirical coverage (observed minus nominal) by lead time over $100$-step rollouts. (c) \gls{vrmse} by lead time. \gls{mc} dropout with MSE is less accurate and reliable overall, despite competitive coverage for \gls{gpe}.}
    \label{fig:ablation_mc_dropout_mse}
\end{figure}

This pattern generally persists over rollouts, with \gls{gpe} the main exception (Figure~\ref{fig:ablation_mc_dropout_mse}). Overall, the baseline is not consistently competitive with either probabilistic training approach.

\clearpage
\subsection{Post-hoc uncertainty calibration}\label{appendix:calibration}

Conformal prediction (CP) turns a heuristic measure of predictive uncertainty into prediction intervals with a finite-sample coverage guarantee~\citep{angelopoulos2021gentle, angelopoulos2026theoretical}. In its split form, it needs only a trained model and a calibration set that the model has not seen: the model's errors on the calibration set, measured relative to its own uncertainty estimate, determine how much its intervals must be widened or narrowed. Conformal prediction has previously been applied to neural surrogates of \glspl{pde}~\citep{gopakumar2026, ma2024}.

We apply split conformal prediction to the $100$-step rollouts of the main \gls{crps} and \gls{fm} models without retraining them. For each dataset, we simulated $150$ additional trajectories with the same \texttt{AutoSim} settings as the datasets of the main results but different random seeds. Rollouts for these trajectories and for the test set of the main results were generated with the checkpoints and evaluation settings of the main results, using ten ensemble members. Of the $150$ new trajectories, $n=100$ ($n=102$ for \gls{gs}, $17$ of each of the six pattern types) form the calibration set, and the remaining $50$ ($48$ for \gls{gs}) are held out for the calibration-set-size check below. All other results are evaluated on the test set of the main results ($20$ trajectories, $24$ for \gls{gs}).

The procedure is applied separately to every output component $r$, i.e.\ every lead time, grid point and channel, as in Section~\ref{sec:coverage_testing}. For calibration trajectory $i=1,\dots,n$, let $x_{i,r}$ be the true value of component $r$, and $\hat{\mu}_{i,r}$ and $\hat{\sigma}_{i,r}$ the mean and standard deviation of the ensemble members. Following \citet{angelopoulos2021gentle}, the score is the error of the ensemble mean in units of the ensemble spread,
\[s_{i,r}=\frac{|x_{i,r}-\hat{\mu}_{i,r}|}{\hat{\sigma}_{i,r}}.\]
For nominal coverage $1-\alpha$, the threshold $\hat{q}_r$ is the $\lceil (n+1)(1-\alpha)\rceil$-th smallest of $s_{1,r},\dots,s_{n,r}$. For a test trajectory with ensemble mean $\hat{\mu}_r$ and standard deviation $\hat{\sigma}_r$, the conformal interval is
\[\bigl[\hat{\mu}_r-\hat{q}_r\hat{\sigma}_r,\;\hat{\mu}_r+\hat{q}_r\hat{\sigma}_r\bigr].\]
The interval is centred on the ensemble mean and proportional to the ensemble spread, and $\hat{q}_r$ sets its width so that a fraction of about $1-\alpha$ of the calibration errors for component $r$ fall inside it. If the calibration and test trajectories are exchangeable, i.e.\ their joint distribution does not change when they are reordered, which holds if both are independent draws from the same simulator settings, the interval contains the truth with probability at least $1-\alpha$, and at most $1-\alpha+1/(n+1)$ when the scores have no ties~\citep{angelopoulos2021gentle}. The guarantee holds for each component marginally, not jointly over the field. It holds for any underlying model, but the intervals are only as informative as the uncertainty estimate they start from: a poor estimate gives wide intervals. We compute the intervals at each nominal level used in the main results, from $0.05$ to $0.95$, and use them in place of the ensemble quantiles when computing empirical coverage (Section~\ref{sec:coverage_testing}) and the Winkler interval score (Section~\ref{section:proper_scoring_rules}). The calibration is implemented in \texttt{AutoUQ}\footnote{\AutoUQRepository}, an open-source Python package for evaluating and calibrating uncertainty estimates.

\begin{figure}[H]
    \centering
    \includegraphics[width=\linewidth]{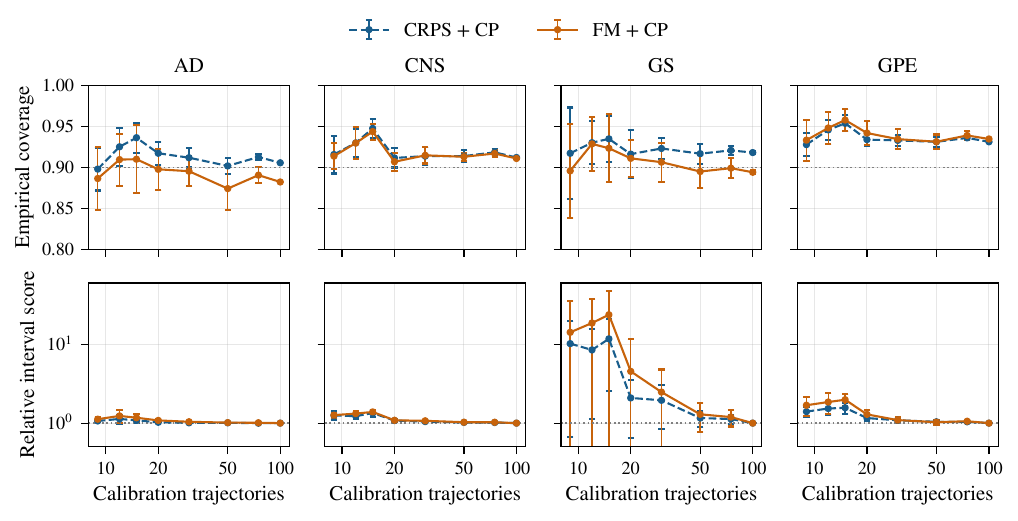}
    \caption{Coverage at $1-\alpha=0.9$ (top) and relative Winkler interval score (bottom) against the number of calibration trajectories $K$, on the held-out new trajectories, averaged over the $100$-step rollout. Mean and standard deviation over $20$ random calibration subsets; bars below the bottom axes are cut. Dashed blue: \gls{crps}; solid orange: \gls{fm}.}
    \label{fig:conformal_calibration_set_size}
\end{figure}

To check that $100$ calibration trajectories are enough, we repeat the calibration on random subsets of $K \in \{9, 12, 15, 20, 30, 50, 75, 100\}$ trajectories drawn from the calibration set, with $20$ draws for each $K$, and evaluate on the remaining $50$ new trajectories ($48$ for \gls{gs}). Figure~\ref{fig:conformal_calibration_set_size} shows 90\% coverage and the Winkler interval score relative to its value at $K=100$, each averaged over the held-out trajectories, grid points, channels and all $100$ lead times, as the mean and standard deviation over the $20$ subsets. Coverage stays between $0.87$ and $0.96$ for every $K$, so the number of calibration trajectories mainly determines how wide the intervals are rather than whether they cover. For $9 \le K \le 18$, $\lceil 0.9(K+1)\rceil=K$, so the threshold is the largest calibration score ($K=9$ is the smallest calibration set for which a 90\% threshold exists) and the interval score reaches up to twice its $K=100$ value on \gls{ad}, \gls{cns} and \gls{gpe}, and $8$--$24$ times on \gls{gs}, where it also varies strongly between subsets, with a standard deviation comparable to or larger than the mean. By $K=50$, the interval score on \gls{ad}, \gls{cns} and \gls{gpe} is within 4\% of its value at $K=100$. On \gls{gs} it is still 13--19\% higher at $K=75$ and has not levelled off, so additional calibration trajectories would narrow the \gls{gs} intervals further, whereas $100$ trajectories are sufficient for the other three datasets. These sizes are specific to these four datasets and are not a general recommendation for how many calibration trajectories conformal prediction requires.

\begin{figure}[H]
    \centering
    \includegraphics[width=\linewidth]{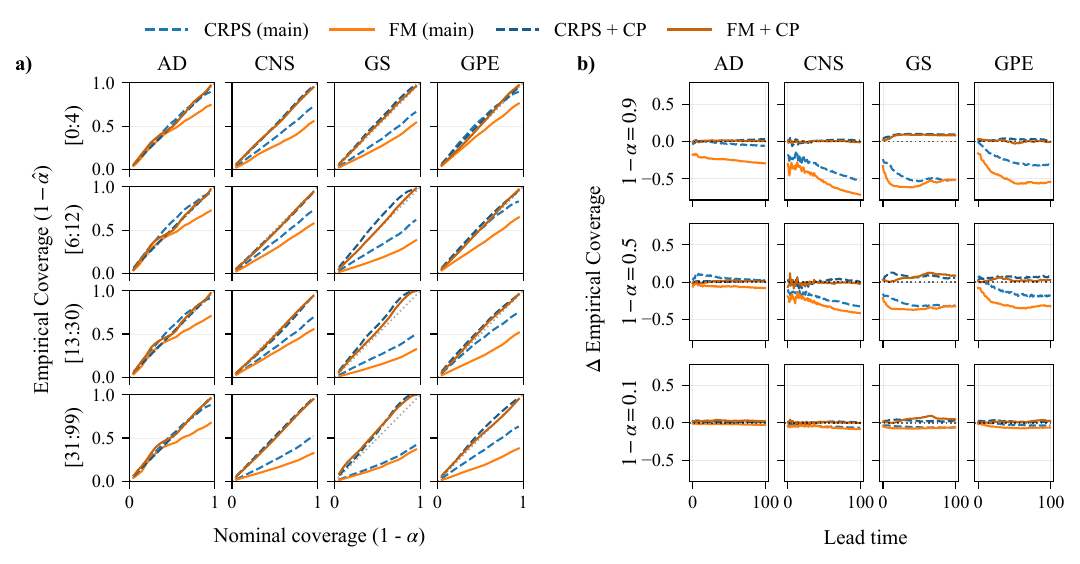}
    \caption{Coverage of the main \gls{crps} (dashed blue) and latent \gls{fm} (solid orange) models before and after conformal prediction (CP; darker shades), fitted on the new calibration set. Each pair uses the same re-evaluated paper test forecasts. (a) Empirical against nominal coverage in each rollout window, as in Figure~\ref{fig:uq_reliability_by_lead_time_panel}. (b) Empirical minus nominal coverage at each lead time for nominal coverage $0.9$, $0.5$ and $0.1$; zero (dotted line) is ideal.}
    \label{fig:calibration_cp_reliability}
\end{figure}

Figure~\ref{fig:calibration_cp_reliability} shows the effect of conformal calibration on the test set. It compares the empirical coverage of the raw ensembles with that of the conformal intervals across nominal levels in each rollout window (a), and shows how far each lies from nominal coverage at every lead time (b). Averaged over test trajectories, grid points, channels and all $100$ lead times, 90\% coverage of the raw ensembles ranges from $0.43$ (\gls{gs}) to $0.87$ (\gls{ad}) for \gls{crps} and from $0.34$ (\gls{gs}) to $0.65$ (\gls{ad}) for \gls{fm}. After conformal calibration it lies between $0.90$ and $0.92$ on \gls{ad}, \gls{cns} and \gls{gpe} for both models, and the Winkler interval score decreases, apart from \gls{crps} on \gls{ad}, where it is unchanged. The shortfall of the raw ensembles grows with lead time: at the 90\% level it widens from $0.17$ below nominal at the first lead time to $0.53$ below at the last for \gls{crps} on \gls{cns}, and from $0.30$ to $0.71$ below for \gls{fm}, with similar growth on \gls{gs} and \gls{gpe}. At the 10\% level, raw coverage stays within $0.09$ of nominal for every model. At the 90\% level, the conformal intervals stay within $0.06$ of nominal at every lead time on \gls{ad}, \gls{cns} and \gls{gpe}. On \gls{gs}, the conformal intervals over-cover the test set ($0.99$ for \gls{crps} and $0.98$ for \gls{fm}), above the upper bound of $0.91$ for $n=102$, whereas the same calibration gives $0.92$ and $0.89$ on the held-out new \gls{gs} trajectories (Figure~\ref{fig:conformal_calibration_set_size}, $K=100$). Per trajectory, most held-out new \gls{gs} trajectories are also covered at $0.96$ or more, but a few ($4$ of $48$ for \gls{crps} and $5$ of $48$ for \gls{fm}, all with maze, spiral or spot patterns) have coverage below $0.8$ and bring the average down to about $0.9$. The $24$ test trajectories include none of these, which at the rate seen on the held-out set happens by chance roughly one time in ten, so the over-coverage is consistent with the small size of the test set (the bound applies to the expected coverage, not to that of one particular set of test trajectories) rather than with a difference between the two sets. The guarantee holds on average over the six parameter settings of \gls{gs}, not for each one: for \gls{crps}, coverage on the held-out trajectories ranges from $0.69$ for maze to $1.00$ for gliders.

\begin{figure}[H]
    \centering
    \includegraphics[width=\linewidth]{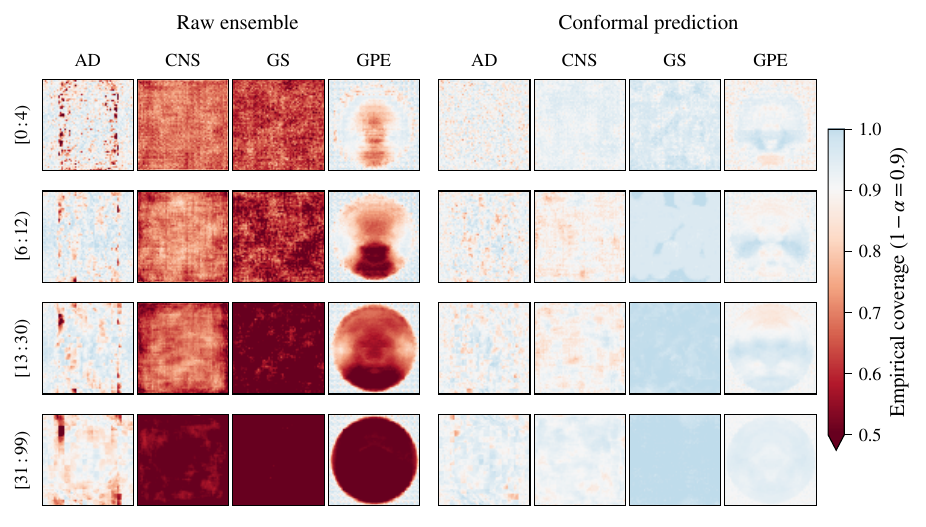}
    \caption{Coverage at $1-\alpha=0.9$ at each grid point for the main \gls{crps} model on the test set, averaged over test trajectories, channels and each rollout window: raw ensemble (left) and conformal prediction (right). The colour scale is linear and white at the nominal $0.9$, with red for under-coverage and blue for over-coverage; values below $0.5$ share the darkest red.}
    \label{fig:conformal_coverage_maps_crps}
\end{figure}

\begin{figure}[H]
    \centering
    \includegraphics[width=\linewidth]{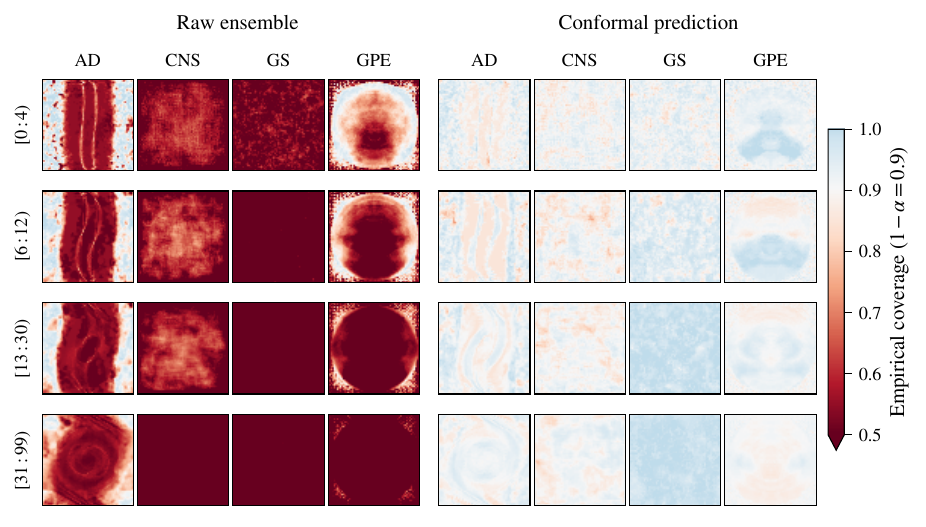}
    \caption{Coverage at $1-\alpha=0.9$ at each grid point for the main latent \gls{fm} model on the test set, averaged over test trajectories, channels and each rollout window: raw ensemble (left) and conformal prediction (right). The colour scale is linear and white at the nominal $0.9$, with red for under-coverage and blue for over-coverage; values below $0.5$ share the darkest red.}
    \label{fig:conformal_coverage_maps_fm}
\end{figure}

Figures~\ref{fig:conformal_coverage_maps_crps} and~\ref{fig:conformal_coverage_maps_fm} show where on the grid the intervals are too narrow or too wide, which coverage averaged over the whole field can hide. Each panel shows 90\% coverage at every grid point on the test set, averaged over test trajectories, channels and the lead times in one rollout window, for the raw ensemble (left four columns) and the conformal intervals (right four columns). With $20$ test trajectories ($24$ for \gls{gs}), differences of a few hundredths between grid points are expected from sampling alone. The raw ensembles under-cover almost everywhere on \gls{cns}, \gls{gs} and \gls{gpe}, increasingly so in later windows, and the \gls{fm} ensemble also under-covers on \gls{ad}. After conformal calibration, most grid points are close to nominal coverage on \gls{ad}, \gls{cns} and \gls{gpe}, with some remaining spatial structure, such as a band of mild over-coverage (about $0.95$) across the centre of the \gls{gpe} domain before the last window. On \gls{gs}, the conformal intervals over-cover across the whole domain, consistent with the field-averaged results above.

\begin{figure}[H]
    \centering
    \includegraphics[width=0.8\linewidth]{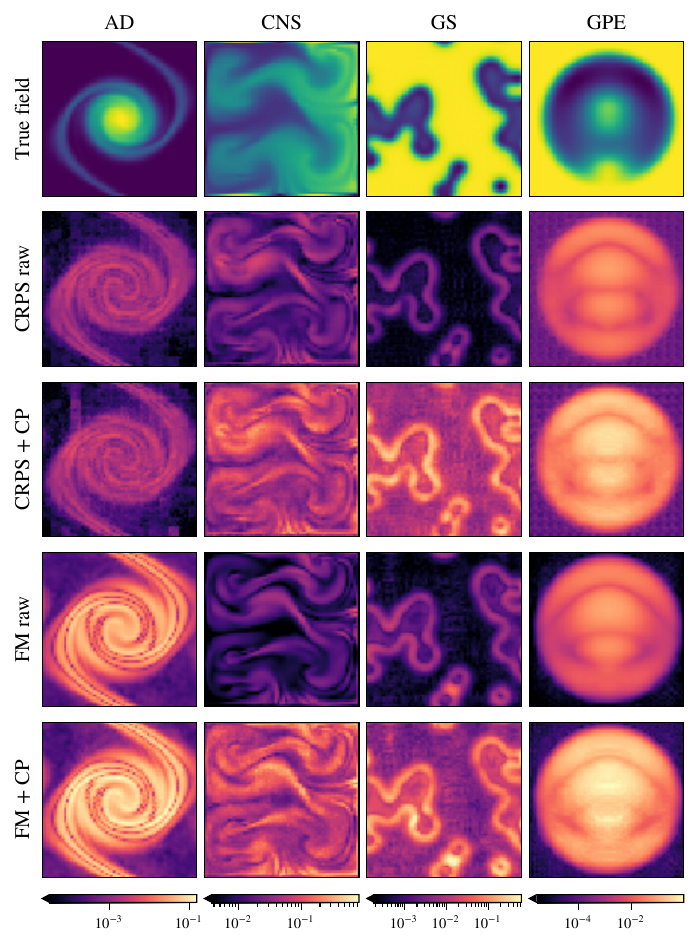}
    \caption{Width of the 90\% intervals at each grid point at lead time $60$ (first channel) for one test trajectory per dataset: the one whose median ratio of conformal to raw width is closest to the median over test trajectories, for both models. Top row: true field. Colour scales are logarithmic and shared within each column.}
    \label{fig:conformal_band_widths}
\end{figure}

Figure~\ref{fig:conformal_band_widths} shows the width of the 90\% intervals at every grid point at lead time $60$ for one representative test trajectory per dataset: the conformal intervals keep the spatial structure of the raw ensemble intervals but are wider. Over the whole test set, taking the ratio of conformal to raw width at every grid point, channel, lead time and trajectory, the median ratio is $1.1$ for \gls{crps} and $1.7$ for \gls{fm} on \gls{ad}, $2.9$ and $6.7$ on \gls{cns}, $3.4$ and $5.9$ on \gls{gpe}, and $18$ and $8.4$ on \gls{gs}, and the ratio grows with lead time. On \gls{gs} and \gls{gpe}, the conformal interval is wider than the range of the true field at 2--8\% of grid points.

\clearpage
\flushbottom

\end{document}